\pdfoutput=1
\documentclass[11pt]{article}
\usepackage[dvipsnames]{xcolor}

\usepackage[preprint]{acl}

\usepackage{times}
\usepackage{latexsym}

\usepackage[T1]{fontenc}
\usepackage[utf8]{inputenc}

\usepackage{microtype}

\usepackage{inconsolata}

\usepackage{graphicx}
\usepackage{tabularx}
\usepackage{multirow}
\usepackage{amsfonts}
\usepackage{xspace}
\usepackage{lipsum}
\usepackage{adjustbox}
\usepackage{booktabs}
\usepackage{makecell}
\usepackage{url}
\usepackage{enumitem}
\usepackage{listings}
\usepackage{xcolor}
\usepackage{inconsolata}
\usepackage[most]{tcolorbox}
\usepackage{caption}
\usepackage{float}
\usepackage{pgfplots}
\pgfplotsset{compat=1.17}
\usepackage{pgfplotstable}

\newcommand{\df}{$\mathbb{DF}$\xspace}
\newcommand{\pf}{$\mathbb{PF}$\xspace}
\newcommand{\dfs}{$\mathbb{DF}$s\xspace}
\newcommand{\pfs}{$\mathbb{PF}$s\xspace}
\newcommand{\dd}{$\mathbb{DD}$\xspace}
\newcommand{\pd}{$\mathbb{PD}$\xspace}
\newcommand{\dds}{$\mathbb{DD}$s\xspace}
\newcommand{\pds}{$\mathbb{PD}$s\xspace}
\newcommand{\bw}{$\texttt{BlocksWorld-100}$\xspace}
\newcommand{\mbw}{$\texttt{MysteryBlocksWorld-100}$\xspace}
\newcommand{\lgt}{$\texttt{Logistics-100}$\xspace}
\newcommand{\bm}{$\texttt{Barman-100}$\xspace}

\title{On the Limit of Language Models as Planning Formalizers}

\author{Cassie Huang \quad Li Zhang \\
  Drexel University \\
  {\tt \{Cassie.Huang\}@drexel.edu} \quad {\tt \{Harry.Zhang\}@drexel.edu}
}

\begin{document}
\maketitle
\begin{abstract}
Large Language Models have been found to create plans that are neither executable nor verifiable in grounded environments. An emerging line of work demonstrates success in using the LLM as a formalizer to generate a formal representation of the planning domain in some language, such as Planning Domain Definition Language (PDDL). This formal representation can be deterministically solved to find a plan. We systematically evaluate this methodology while bridging some major gaps. While previous work only generates a partial PDDL representation, given templated, and therefore unrealistic environment descriptions, we generate the complete representation given descriptions of various naturalness levels. Among an array of observations critical to improve LLMs' formal planning abilities, we note that most large enough models can effectively formalize descriptions as PDDL, outperforming those directly generating plans, while being robust to lexical perturbation. As the descriptions become more natural-sounding, we observe a decrease in performance and provide detailed error analysis.\footnote{Our code and data can be found at \url{https://github.com/CassieHuang22/llm-as-pddl-formalizer}.}
%\footnote{Our code and data can be found at \url{anonymous.4open.science/r/llm-as-pddl-formalizer-1BE2}.}
\end{abstract}

% This work is one step further to using LLMs for formal planning and gives insight on where models fall short and where to improve from there.
% Maybe we don't need this

\section{Introduction}

%The use of Large Language Models has sky-rocketed in the past few years with the performance of GPT far surpassing previous baselines. This performance has led to much research on the use of LLMs for text-based, automated planning and reasoning tasks. Previous work has seen giving the LLM the problem, and asking it to directly generate the plan. However, performance falls short due to the lack of reasoning capabilities in the LLM. 

%\hc{Re: Cassie's version. The 1st sentence may be unnecessary since the claim is obvious nowadays. Sentence 2 and 3 nicely lay out the problem, but would likely benefit from some examples for lay-readers. Sentence 4: always back up a claim! How do we know LLMs lack reasoning ability? Citations!}

%\hc{Describe gap with examples} 
Large Language Models (LLMs) can make \textit{informal plans}, such as suggesting ideas for parties or giving general advice on immigration. However, most users, let alone automated agents like robots, would not be able to actually execute those plans step-by-step to fruition -- either to organize parties or acquire visas -- without significant prior knowledge or external help. This inability to make executable plans lies in LLMs' inability of grounding and formal reasoning \cite{liu2023evaluating,pan-etal-2023-logic,zhang-etal-2023-causal}. 
%\hc{Describe the first attempt on this, and why it's insufficient} 
Cutting-edge research in the community has evaluated LLMs' ability to make \textit{formal plans} in grounded environments, such as textual simulations, where all objects and actions represent actualities in the real world. Therefore, any resulting plan that formally involves those objects and actions would be executable and verifiable by nature. Although formal planning has been desirable in the history of AI \cite{weld1999recent}, recent work found that even state-of-the-art LLMs are unable to generate formal plans \cite{silver2024generalized,valmeekam2024planbench,stechly2024chain}.

%\hc{The biggest change I propose in my version above is to try to get people excited. Give examples. Threaten with consequences (user cannot execute plans if LLMs are as is). Give promises (plans should be executable and explainable). Also, cite a couple of seminal works to corroborate key claims.}

\begin{figure*}[!ht]
    \centering
    \includegraphics[width=\textwidth]{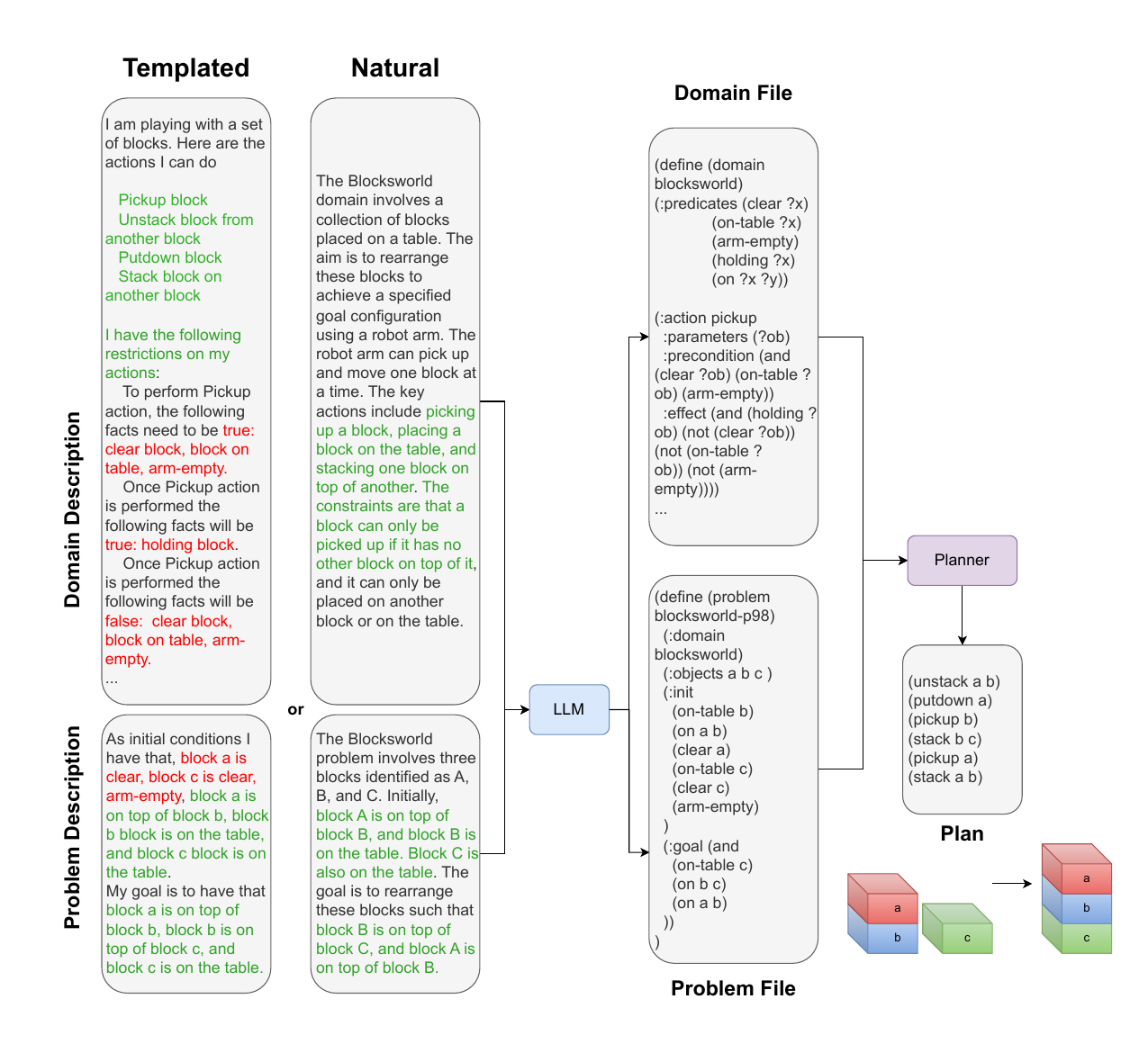}
    \caption{LLM-as-formalizer uses natural language descriptions to generate the Domain and Problem File in PDDL, then these are given to a planner to find a plan. We explore the effect of natural-ness of the language in the description, by giving the model both templated and natural descriptions. Examples of Domain Descriptions and Problem Descriptions from the BlocksWorld Domain are shown. The \textcolor{OliveGreen}{green text} displays what the two examples have in common (listing all possible actions and restrictions) and the \textcolor{red}{red text} displays text that is not considered natural. The ``Templated'' text corresponds to the ``Heavily Templated'' version discussed in Section~\ref{sec:evaluation}. }
    \label{fig:llm as formalizer}
\end{figure*}

Instead of using the LLM as a planner to generate the plan directly, an alternative line of work uses the LLM as a \textit{formalizer}. 
%\hc{Let's use the word "formal" instead of "symbolic" throughout to reduce terminologies.} 
Here, the LLM generates a formal representation of a planning domain, for example in PDDL, based on some natural language descriptions of the environment. This representation can then be fed into a solver to find the plan deterministically (see Figure~\ref{fig:llm as formalizer}). 
%\hc{There are many ways to describe a methodology, but only one optimal way exists which is the most precise and succinct. I'm not sure what it is either, but we'll try.}
Previous work has shown feasibility and success of LLM-as-formalizer over LLM-as-planner in various domains \cite{lyu-etal-2023-faithful,xie2023translating,liu2023llm+,zhang-etal-2024-pddlego,zhang-etal-2024-proc2pddl,zhu2024language}, as LLMs are more capable of information extraction than formal reasoning \cite{zhang-etal-2024-openpi2}. 
%\hc{Describe the two gaps} 
However, the above work has two major shortcomings. First, to simplify the task and evaluation, most have only attempted to generate a partial PDDL representation while assuming the rest is provided, which is often unrealistic in real life. Second, the language used to describe the environments is often artificially templated and structured, leading to potential overestimation of models' ability.

This paper explores the limit of LLM-as-formalizer devoid of the above two simplifications. We use LLMs to generate the entirety of a PDDL representation, including the domain file and the problem file, given a natural-sounding description of the environment and the task (see Figure~\ref{fig:llm as formalizer}). On four widely used planning simulations from the International Planning Competition, we benchmark a suite of LLMs on generating PDDL that is both solvable and correct. As the descriptions in these datasets are templated, we also construct model-generated, human-validated descriptions that are natural-sounding to different levels. 

%The input to the LLM is a problem description written in natural language. The problem can be natural sounding and can come up in real life, or can be templated and sounds like an English version of PDDL. We ask the LLM to translate the problem into a domain and problem file in PDDL which we then give to the solver to find the plan.

Our work is the first to systematically analyze state-of-the-art LLMs' ability to use the trending methodology of LLM-as-formalizer on the highly challenging task of formal planning. We put forward an array of observations that will benefit future efforts. Discussed in detail in Section~\ref{sec:results}, our key findings are as follows.\begin{itemize}[topsep=0pt,itemsep=-1ex,partopsep=1ex,parsep=1ex]
    \item Strongest models like \texttt{gpt-4o}, \texttt{o3-mini}, or \texttt{DeepSeek-R1} can not only plan but also formalize entire PDDL, while other models like \texttt{Gemma-2} or \texttt{Llama-3.1} (up to 405B) cannot do either.
    \item LLM-as-formalizer often outperforms LLM-as-planner for many model-data combinations, though not always.
    \item A more human-like, natural description of the planning environment creates more challenges for models.
    \item LLM-as-formalizer is much more robust to long-tail lexical distribution than LLM-as-planner. 
    \item Weaker models commonly succumb to syntax errors, while all models face semantic errors.
\end{itemize}

%We show that LLMs can generate the entirety of PDDL to different extent in a few-shot setting.
%For example, on the original, templated BlocksWorld dataset, a large model GPT4o-as-formalizer achieves 89\% correctness, greatly outperforming the as-planner counterpart with 29\%; a smaller model GPT4o-mini shows a similar trend of 58\% over 0\%. 
%However, as the descriptions sound more natural, the task becomes more challenging, where the performance of GPT4o-as-formalizer drops to 59\%, while still trailing the as-planner counterpart with 33\%.
%Next, open-source models such as Llama are unable to generate well-formed PDDL. Well-performing models, on the other hand, pass the 'wug test' when the names of entities and actions are obfuscated, disproving the possibility of regurgitation. We also provide fine-grained manual error analyses on all datasets and possible reasons for these errors. 

\section{Task: Formal Planning with PDDL}

\iffalse
\begin{figure}
    \centering
    \includegraphics[width=\columnwidth]{figures/pddl.png}
    \caption{Components of PDDL. A PDDL solver produces a plan from a domain file and problem file. Previous work assumes that portions of the file, or the entire file is given, while we predict the entire PDDL.}
    \label{fig:pddl}
\end{figure}
\fi

Formal planning (or classical planning) with PDDL involves a domain file (\df) and problem file (\pf) (Figure~\ref{fig:llm as formalizer}). The \df describes general properties in a planning domain that hold true across problems, while the \pf describes specific configurations of each problem instance. Concretely, the \df defines all available actions for a state-based environment as well as predicates that represent the properties of object types. Each action definition contains the name of the action, parameters and semantics. The semantics of an action includes the preconditions which describe the necessary world states where the action is valid to execute, and effects which describe how the states change after the action is executed.
The \pf defines the involved objects, initial states, and goal states. These two files are then given to a deterministic planner which will algorithmically search for a plan. Such a plan is a series of executable, instantiated actions that sequentially transition the world states from initial to goal. In the AI community, classical planning has been deemed an effective approach to solve real-world users' problems, as the process is precise, explainable, verifiable, and deterministic.

However, formal planning demands a well-crafted \df and \pf pair. In a real-world planning scenario, an average user would not describe their environment and problem with PDDL, but more likely with a textual domain description (\dd) and problem description (\pd). %, which can be written in detail or omit some information. 
While the \dd and \pd technically describe all the necessary information from the \df and \pf in natural language, information that can be easily inferred by humans is omitted as the language in the descriptions gets more natural. For example, when describing the action ``picking up a block'', a very natural \dd would omit the effect that our hand is no longer empty. %The behavior will not be written in the \dd. %The \pd describes the initial state and goal state of the problem in natural language. 
These descriptions are then translated into PDDL, which can then be used to find a plan when given to a solver.
Hence, we focus on the textual flavor of formal planning: given a \dd and \pd, the model outputs a successful plan with regard to the \df and \pf that are withheld from the model. 

%1 page.

%Refer to Zhu et al Section 3, Guan et al Section 3, PDDLego Section 2. Should discuss \df, \pf, \dd, \pd, plan, solver, etc. 

%There could be a Figure 2 to illustrate the above components (plenty of references in related work, and also my slides). In contrast, Figure 1 would illustrate different natural-level of descriptions, but no need to go into details of \df and \pf. 

\section{Methodology: LLM-as-Formalizer}

\begin{figure}[!t]
    \centering
    \includegraphics[width=\columnwidth]{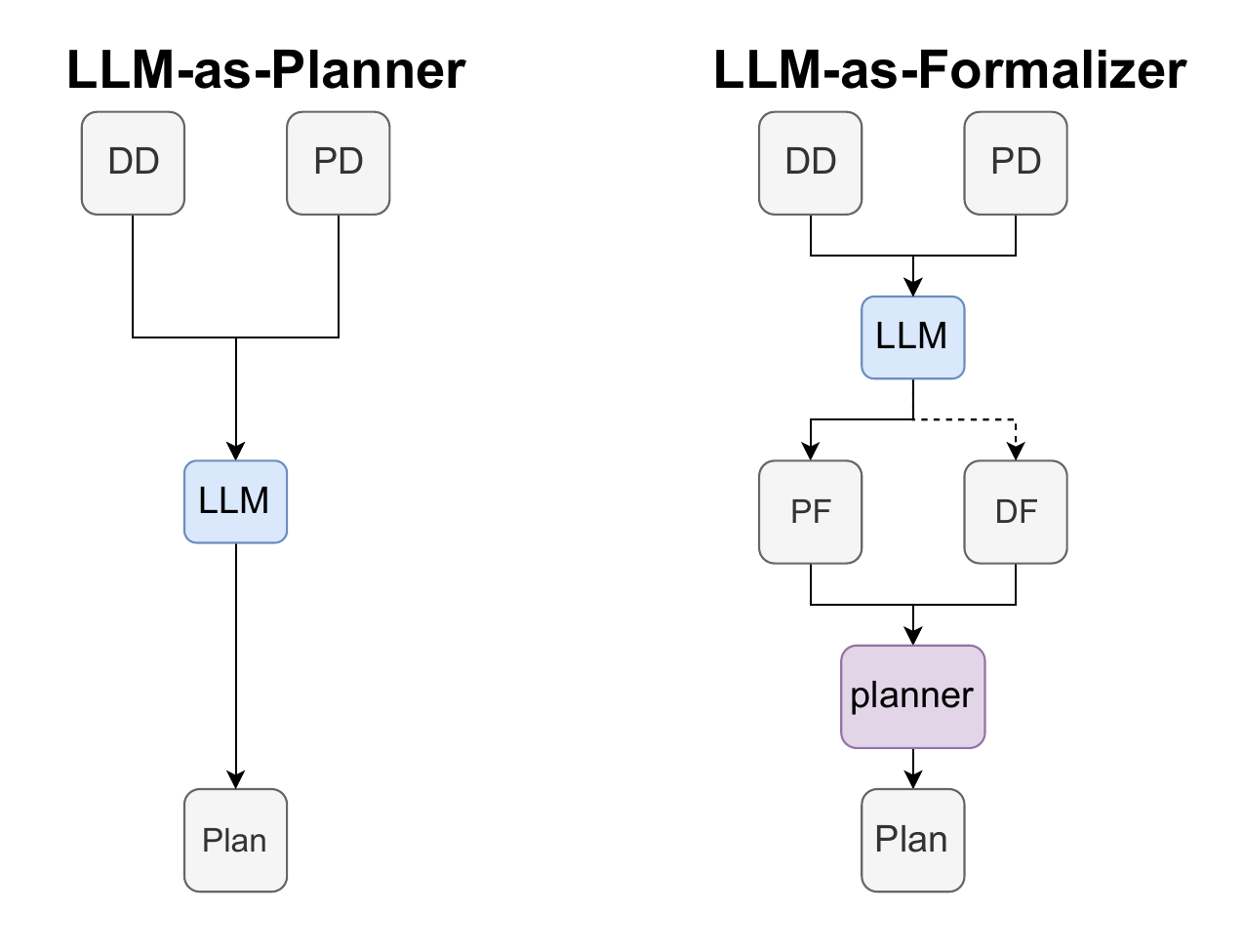}
    \caption{Methodologies for using LLMs in planning. LLM-as-Planner generates the plan directly, while LLM-as-Formalizer translates the \dd and \pd into PDDL. Previous work like \citet{liu2023llm+} use the LLM to generate partial PDDL such as \pf only, while we generate the entire PDDL including  \pf and \df. Note the \dd and \pd are always provided and the \df and \pf are always generated by the LLM.}
    \label{fig:two_methodologies}
\end{figure}

To tackle the task above, recent work involving LLMs diverged into two methodologies. The first, \textbf{LLM-as-planner}, directly uses LLMs to generate a plan based on the \dd and \pd. The second, \textbf{LLM-as-formalizer}, uses LLMs to recover the \df and \pf, before using a deterministic planner to arrive at the plan (Figure~\ref{fig:two_methodologies}). Our work will focus on the second while using the first as a baseline. LLM-as-formalizer is in essence neuro-symbolic, where LLMs help define the structured representation that is otherwise prohibitively costly to annotate and brittle to generalize. 
%If a plan is found, we verify the correctness of the plan by again using a simulator.
%For the LLM-as-planner approach, we give the LLM a \dd and a \pd, and ask it output a plan. In addition, we provide LLMs with the action space as well as an example plan, so that the model is given all necessary information to output a structured plan. With the new ability of LLMs to work with code, previous work has also attempted to provide LLMs with \df and \pf, so the LLM's job was exactly the same as a traditional planner, with unsatisfactory performance. In our work, we consider this intuitive method a baseline.
%The plans are verified using a simulator. 
%Harry: talk about this later in Sec 4
%For the LLM-as-formalizer approach, the input is also the \dd and \pd. However, instead of directly outputting the plan, the LLM outputs PDDL which includes the \df and \pf. These two files are then fed to a search-based planner to arrive at a deterministic plan. 
%Harry: talk about this later in Sec 4
Existing works in this line demonstrated success but only generate a partial PDDL representation, while assuming the rest. For example, some works generate only the \pf goals while the \df and remaining \pf are given \cite{lyu-etal-2023-faithful,xie2023translating}. Others generate the entire \pf while the \df is given \cite{liu2023llm+,zhang-etal-2024-pddlego,zuo2024planetarium}. There have also been works that generate the action semantics in the \df while the remaining parts of the \df and entire \pf are provided \cite{zhang-etal-2024-proc2pddl,zhu2024language}, and work that generates the \df while the \pf is given \cite{wong2023learning,guan2023leveraging}. While these assumptions of given PDDL components simplify the task and evaluation, they are often unrealistic. Therefore, we bridge this gap by asking the LLM to predict the entire PDDL -- both the \df and \pf.\footnote{It is however minimally necessary to provide the action space, the identifiers and parameters of the actions in \df, so the agent knows \textit{what} actions are possible.}

%0.5 page.

%Start with the LLM-as-planner baseline, then focus on LLM-as-formalizer, e.g., what their input and output is. Also compare with related work to demonstrate how they did not predict the entire PDDL. 

\section{Evaluation: Datasets, Metrics, Models}
\label{sec:evaluation}

To evaluate both approaches above, we work with \textit{fully-observed} textual environments. Here, the provided \dd and \pd contain all necessary information for the model to make a complete plan.
%Therefore, the model is expected generate the entire plan without needing to interact with the environment as it would in \textit{partially-observed} environments.

\subsection{Datasets}
We consider four simulated planning domains, BlocksWorld, Logistics, Barman from the International Planning Competition \cite{ipc}, and Mystery BlocksWorld \cite{valmeekam2024planbench}.

\noindent\textbf{BlocksWorld}, also used in \citet{liu2023llm+}, is a domain to rearrange stacks of blocks on a table using a robotic arm. The domain includes 1 type of entity, 5 predicates, and 4 actions.\\
\noindent \textbf{Mystery BlocksWorld} obfuscates the original BlocksWorld domain by replacing all the names of the types, predicates, actions, and objects with nonsensical words, akin to a \textit{wug test} \cite{berko1958child}. This dataset as a control group can effectively test whether models create plans via lexical pattern-matching and memorization.\\ 
%The original Mystery BlocksWorld dataset is Heavily Templated. Since the dataset itself is an artificial perturbation, we do not provide more natural descriptions as that would defeat its purpose. We sampled 100 random problems for our experiment, resulting in 1 \df, 1 \dd, 100 \pfs, and 100 \pds. We refer to this dataset as \mbw.
\noindent \textbf{Logistics}, also used in \citet{guan2023leveraging}, is a domain to transport packages across different locations using both trucks and airplanes. In this domain, there are 6 types of entities, 3 predicates, and 6 actions.\\
\noindent \textbf{Barman}, also used in \citet{zhu2024language}, is a domain to create cocktails from ingredients using different containers and two robotic arms. In this domain, there are 7 types of entities, 13 predicates, and 12 actions. 

Each dataset comes with ground-truth PDDL files describing the domain (\df) and problem (\pf). The input to the model is a natural language description of the domain (\dd) and problem (\pd). The output of the model is a plan, namely a sequence of instantiated actions defined in the \df. For each of these datasets, the natural language description (\dds and \pds) were created in 3 different levels of naturalness.

\noindent\textbf{Heavily Templated}. For BlocksWorld, Logistics and 
Barman, the Heavily Templated \dd and \pd are generated using the same template as Mystery BlocksWorld \cite{valmeekam2024planbench}. This description is almost a word-by-word translation of PDDL. For example, for the `pick-up' action in BlocksWorld, the ground-truth PDDL \df would be the following:

\resizebox{0.9\linewidth}{!}{\begin{tcolorbox} [top=2pt,bottom=2pt, width=\linewidth, boxrule=1pt]
    {\small {\fontfamily{zi4}\selectfont
    (:action pick-up \\
    :parameters (?b - block) \\
    :precondition (and (clear ?b) (on-table ?b) (arm-empty)) \\
    :effect (and (not (on-table ?b)) (not (clear ?b)) (not (arm-empty)) (holding ?b)) \\
  ) 
    }
    \par}
\end{tcolorbox}}

while the Heavily Templated \dd is: 

\resizebox{0.9\linewidth}{!}{\begin{tcolorbox} [top=2pt,bottom=2pt, width=\linewidth, boxrule=1pt]
    {\small {\fontfamily{zi4}\selectfont
    To perform Pickup action, the following facts need to be true: clear block, block on table, arm-empty. \\
    Once Pickup action is performed the following facts will be true: holding block. \\
    Once Pickup action is performed the following facts will be false:  clear block, block on table, arm-empty.
    }
    \par}
\end{tcolorbox}}
    
From an application point of view, spelling out all preconditions and effects in terms of the predicates is paradoxical, as it assumes the user already has the algorithmic awareness of PDDL.

%\hc{TODO: give a unified example across these three naturalness levels; perhaps the effect of pick-up}
%and uses the name of predicates instead of writing them out naturally (eg. "arm-empty" vs. "the arm is empty").

\noindent\textbf{Moderately Templated}. The \dd and \pd are generated using the same template as the original BlocksWorld dataset, following \citet{valmeekam2024planbench}. For example, the Moderately Templated description of the `pick-up' action is:

\resizebox{0.9\linewidth}{!}{\begin{tcolorbox} [top=2pt,bottom=2pt, width=\linewidth, boxrule=1pt]
    {\small {\fontfamily{zi4}\selectfont
    I can only pick up or unstack one block at a time. \\
   I can only pick up or unstack a block if my hand is empty. \\
   I can only pick up a block if the block is clear. A block is clear if the block has no other blocks on top of it and if the block is not picked up.
    }
    \par}
\end{tcolorbox}}
While more natural-sounding than the Heavily Templated version, the description still explicitly discusses the preconditions and effects as well as predicates like `clear'. 
%\hc{TODO: ditto}
%(eg. "block a is clear", but a human would not need to be told a block is clear because it is already at the top of a stack and has no other block on top).

\noindent\textbf{Natural}. A realistic pair of \dd and \pd should emulate how real-life users would describe the planning domain and problem, such that a human problem solver would understand and have just enough information to find a plan. To create such descriptions, we use a human-in-the-loop, model-assisted data generation approach. 

To generate \dds, we ask \texttt{gpt-4o} with high temperature to generate and paraphrase a seed annotated \dd, and then manually verify the correctness by making sure it lists the correct predicates, preconditions and effects, which are not unique. We next verify the naturalness of the generated text by making sure there were variations in language throughout all descriptions generated, but unnecessary information was not given out. 
%{\color{purple}{For the Logistics domain, we generated the \dd by asking GPT-4o to humanize the Moderately Templated \dd generated previously.}}
%For example, we remove descriptions talking about the size of the blocks which is irrelevant to solving the task. 

To generate \pds, we provide the model with a symbolic configuration containing the initial and goal states of the problem. The model then `humanizes' the problem by making it sound natural, given a couple of seed exemplars. We manually verify the correctness of the datasets of the non-templated problems by hand by comparing them against the problem configurations. We then verify the naturalness of the \pd by making sure there is variation but no ambiguity in its language.
%{\color{purple}{ For the Logistics domain, we generated the \pd by asking GPT-4o to humanize the Moderately Templated \pd generated previously, then we manually verify for accuracy.}}

%we first attempted to ask the model to generate the problem at will. The model was given the total number of blocks and generated its own problems. However, these problems are often inconsistent with regard to the number and type of the blocks. Therefore, 

\begin{figure*}[t!]
    \centering
    \includegraphics[width=\linewidth]{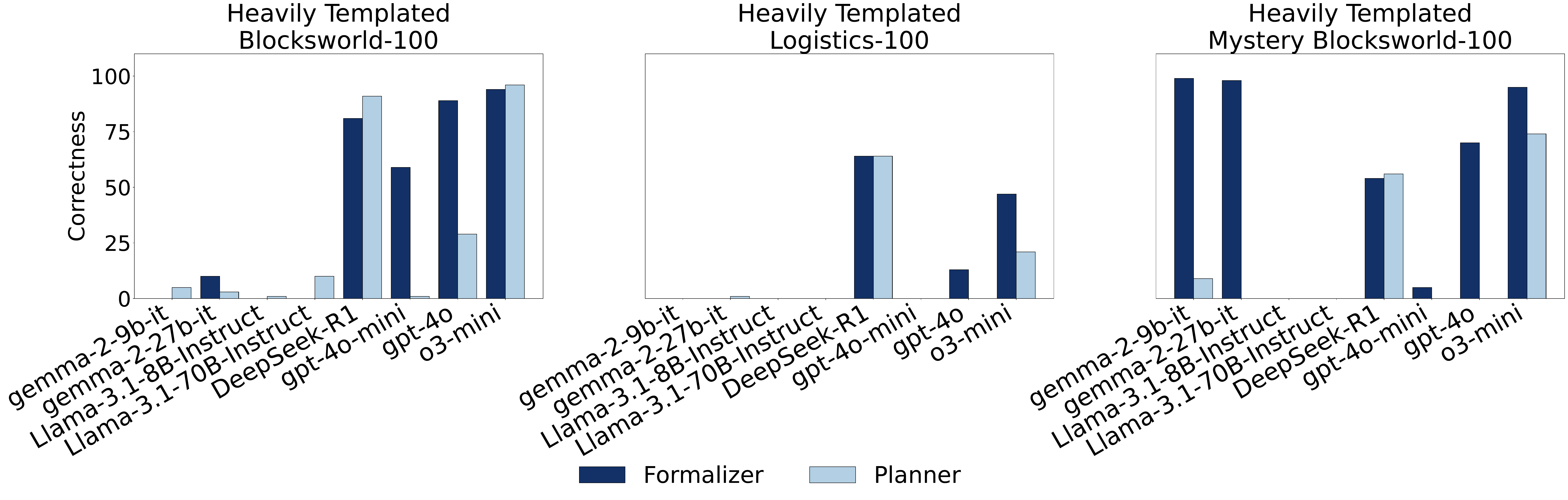}
    \caption{Performance of both usages of LLMs on Heavily Templated \bw, \lgt, and \mbw. Detailed results are shown in Appendix~\ref{app:detailed_results}. Due to the cost and zero solvability on our easiest dataset, results of \texttt{Llama-405B} and \texttt{DeepSeek-8B|70B} are omitted.}
    \label{fig:all_results}
\end{figure*}

In the following example of a Natural description of BlocksWorld:

\resizebox{0.9\linewidth}{!}{\begin{tcolorbox} [top=2pt,bottom=2pt, width=\linewidth, boxrule=1pt]
    {\small {\fontfamily{zi4}\selectfont
    The robot arm can pick up and move one block at a time from one position to another. It is only able to move the top block from any stack or table, and have only one block held by the robot arm at a time. The main actions available are 'pick up', ...
    }
    \par}
\end{tcolorbox}}

The example no longer discusses the preconditions and effects of each action one by one, but rather focuses on the general rules to the domain. These rules apply to not only `pick-up' but also other actions. Therefore, the \dd can be much more concise, requires less algorithmic awareness, and is more realistic.

In total, we construct 100 problems varying in complexity for all domains. For each of the two Templated descriptions, there is one \dd paired with each of 100 \pds. For the Natural descriptions, there are 100 different pairs of \dds and \pds. We refer to these datasets as \texttt{BlocksWorld-100}, \mbw, \texttt{Logistics-100} and \texttt{Barman-100}. Moderately Templated and Natural datasets are available only for \texttt{BlocksWorld-100} and \texttt{Logistics-100} due to the size of \texttt{Barman-100}. Data examples can be found in Appendix~\ref{sec:data_examples}.

\subsection{Metrics}
Following past work \cite{guan2023leveraging,zhu2024language}, the plan produced from LLM-as-Planner is validated using VAL \cite{1374201} against the ground-truth \dfs and \pfs provided above. This is done instead of being compared against ``ground-truth'' plans like some work \cite{lyu-etal-2023-faithful,liu2023evaluating,pan-etal-2023-logic} since there could be multiple correct plans. Similarly, the predicted \df and \pf for the LLM-as-formalizer approach are not compared against the ground-truth, as only the eventual plan is validated because there might be more than one way to formalize the planning domain and problem in PDDL. 

We evaluate the predicted plans following metrics from \citet{zuo2024planetarium}: using \textit{solvability} and \textit{correctness}. Solvability only applies to LLM-as-formalizer and indicates the percentage of solvable \textit{predicted} \dfs and \pfs, regardless of whether the resulting plan can be executed based on the \textit{ground-truth} \df and \pf. Correctness indicates the percentage of actually correct plans. However, different from \citet{zuo2024planetarium}, Solvability was determined using the planner \texttt{dual-bfws-ffparser} implemented by \citet{muise-icaps16demo-pd} and Correctness was evaluated using VAL\footnote{\url{nms.kcl.ac.uk/planning/software/val.html}}. More information about the solver and validator can be found in the Appendix, Section \ref{sec:solver info}.

\subsection{Models} \label{models}

For both of the LLM-as-planner and LLM-as-formalizer approaches, we consider a number of models, including open-source and closed-source LLMs varying in size, including gemma-2-9b|27b-it \cite{team2024gemma}, llama-3.1-8B|70B|405B-Instruct \cite{dubey2024llama}, DeepSeek-R1-Distill-Llama-8B|70B|671B \cite{guo2025deepseek}, gpt-4o-mini-2024-07-18, gpt-4o-2024-08-06, and o3-mini-2025-01-31\footnote{\url{platform.openai.com/docs/models}}. We query these models using KANI \cite{zhu-etal-2023-kani} with default hyper-parameters. The open-source models are run using 4 RTX A6000 GPUs, averaging about 1062 input and output tokens for the LLM-as-formalizer approach in \bw. 

To emulate real-life application with minimal user interference, we use zero-shot prompts for all naturalness levels across all datasets (see prompts in Appendix~\ref{sec:prompts}). We would also like to note that since the LLM has to generate the entire \df and \pf, using few-shot prompting would not be meaningful. Each dataset corresponds to one unique domain and we cannot provide the LLM the \df in the prompt as the model is supposed to predict it. As for the \pf, we also cannot provide it to the model because the typing, predicates, and usage in the \pf would also give away critical information from the file. Nevertheless, we performed a small set of experiments using different prompting methods, such as few-shot and chain-of-thought prompting on Heavily Templated and Natural \bw. Results and discussion of these experiments can be found in the Appendix Sections \ref{few-shot} and \ref{chain-of-thought}.

\section{Results and Observations}
\label{sec:results}

In this section, we display our results as well as perform an in-depth analysis of the strengths and weaknesses of LLMs in formal planning, to understand the impact of the model choice, naturalness of the description, content of the task, and difficulty of the problem.

\subsection{Can LLMs formalize?}

We seek to understand the extent to which LLMs can act as a formalizer to generate \textit{entire} PDDL, rather than partial components seen in previous work. Figure \ref{fig:all_results} displays the results of our experiments on Heavily Templated  \bw, \lgt and \mbw. Results on the most complex domain \bm is omitted due to close-to-zero performance for all models.

These results demonstrate that \textbf{GPT-family LLMs and \texttt{DeepSeek-R1} can decently generate PDDL}, while open-source models like Gemma and Llama (even up to 405B parameters) struggle. As a formalizer, \texttt{gpt-4o-mini}, \texttt{gpt-4o} and \texttt{o3-mini} demonstrate non-trivial and increasing performance on \bw. On the more complex \lgt, \texttt{gpt-4o-mini} succumbs to zero performance, whereas the other two show decreased performance. The solvability of \texttt{gpt-4o-mini} is often much higher than its correctness, suggesting a good grasp of the PDDL syntax but a lack of semantic understanding. In contrast, the solvability of \texttt{gpt-4o} and \texttt{o3-mini} is often 80\% to 100\% of their correctness. On the other hand, \textbf{open-sourced models can rarely generate PDDL}, a low-resource language, despite them being reportedly strong at generating high-resource languages like Python \cite{cassano2022multipl}. All Llama models up to 405B cannot generate any solvable PDDL across all three datasets, while Gemma models show poor, though non-zero, performance on \bw and \lgt, and strong performance on \mbw. 

\subsection{Should LLMs formalize?}
Between LLM-as-planner and LLM-as-formalizer, which is the preferred methodology? Figure~\ref{fig:all_results} shows that on \bw, \texttt{gpt-4o} is able to generate solvable PDDL 64 times, out of 100, and of those 64 plans, 60 of them are correct. This far surpasses the LLM-as-planner baseline, which only found correct plans 33/100 times. This trend holds for \lgt as well as the Moderately Templated and Natural \bw data (Figure~\ref{fig:results_bw_across_naturalness}). On \mbw, we can see that LLM-as-formalizer can generate 70/100 correct plans, which far surpassed LLM-as-planner which did not find a single correct plan as the description becomes unorthodox. The superiority of LLM-as-formalizer also extends to \texttt{gpt-4o-mini} but not \texttt{o3-mini}, who shows strong performance as a planner. 
These results demonstrate that \textbf{LLM-as-formalizer greatly outperforms LLM-as-planner} in most cases, whenever these LLMs can formalize PDDL \textit{at all}. However, these results also show that models that cannot formalize (e.g., Llama models) can still plan, though with close-to-zero performance.

\subsection{The more natural, the harder?} 
\label{sec:analysis_naturalness}

\begin{figure*}[t!]
    \centering
    \includegraphics[width=\linewidth]{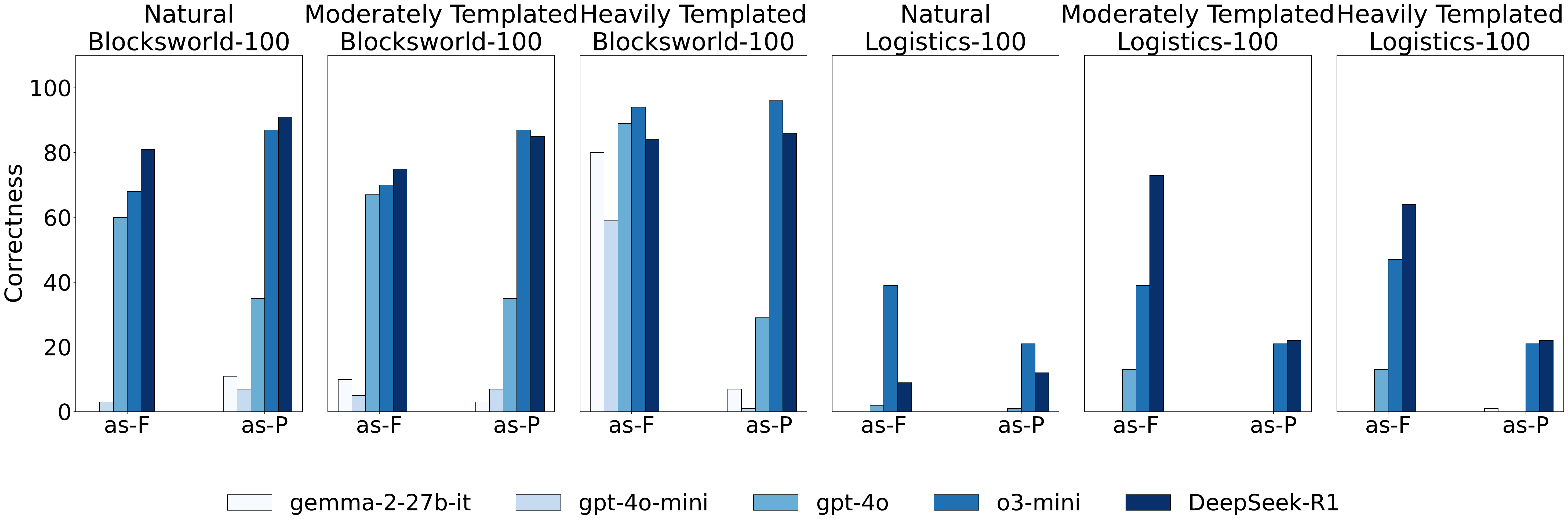}
    \caption{Performance of LLM-as-planner (as-P) and LLM-as-formalizer (as-F) across different naturalness level of description on \bw and \lgt. Detailed results are shown in Appendix~\ref{app:detailed_results}. }
    \label{fig:results_bw_across_naturalness}
\end{figure*}
We now examine whether using humanized descriptions makes the problem more difficult. Results from Figure~\ref{fig:results_bw_across_naturalness} show that on \bw as the problem sounds more similar to PDDL and less natural, the performance of all the models improves. Similar results hold for the Logistics domain (see results in Appendix~\ref{app:detailed_results}). This suggests that \textbf{a more natural-sounding domain and problem description is much more challenging} than templated, less natural sounding descriptions. One potential explanation is that pattern matching a template back to PDDL is much easier than having to first parse all the predicates and objects from a passage. Another reason is a more natural sounding description may leave out implicit common-sense. For example, the Natural \bw data does not explicitly specify that a block is `clear', because any human who reads that a block is ``on top of a stack'' can understand that there is no block on top of it and hence `clear' to be moved. However, models often fail to invoke this knowledge and will leave out the `clear' predicate, leading to unsolvable PDDL or incorrect plans.

\subsection{Do LLMs memorize pretraining?} \label{mbw}
\iffalse
\begin{figure}
    \centering
    \includegraphics[width=\columnwidth]{figures/memorization.pdf}
    \caption{Performance of Natural \bw versus Heavily Templated \mbw.}
    \label{fig:mbw}
\end{figure}
\fi
\iffalse
\begin{table}[t!]
    \centering
    \small
    \begin{tabular}{lll} \toprule
    %& \multicolumn{2}{c}{Metrics} \\ \midrule
      Models & Solvability & Correctness \\ \midrule
    %\multirow{3}{4em}{Small} & \texttt{gpt-3.5-turbo} & 4/100 & 0/100 \\
    % & \texttt{Llama-3.1-8B} & 0/100 & 0/100\\
    % & \texttt{gemma-2-9b-it} & 100/100 & 99/100 \\ \midrule
    % \multirow{3}{4em}{Medium} & \texttt{gpt-4o-mini} & 36/100 & 5/100 \\ 
    % & \texttt{Llama-3.1-70B} & 0/100 & 0/100 \\
    % & \texttt{gemma-2-27b-it} & 99/100 & 98/100 \\ 
      \texttt{gpt-4o-mini} & 36/100 & 5/100 \\
     \texttt{gpt-4o-mini}$^{p}$ & - & 0/100 \\
      \texttt{gpt-4o} & 74/100 & 70/100 \\ 
     \texttt{gpt-4o}$^{p}$ & - & 0/100 \\
\bottomrule     
    \end{tabular}
    \caption{Performance of LLM-as-formalizer and LLM-as-planner ($^{p}$) on \mbw.}
    \label{tab:Heavily Templated mbw}
\end{table}
\fi

Do LLMs generate plans or formalize PDDL based on what they have memorized in their training data? We determine this by looking at the results on \mbw, a derivative of BlocksWorld where all names are perturbed and nonsensical. From Figure~\ref{fig:all_results}, we can see that LLM-as-planner was not able to find a single correct plan using either \texttt{gpt-4o-mini} or \texttt{gpt-4o}. However, \texttt{gpt-4o}-as-formalizer surpassed this baseline with a Correctness score of 70/100. Surprisingly, Gemma models performed very well on the \mbw dataset with near perfect solvability and correctness for both models. We conclude that the results are mixed across models, which contradicts previous works cited involving PDDL generation which seem to claim that LLM-as-formalizer is clearly better than other methodologies. Despite the inconclusiveness among models, there are more clear-cut results between two groups of models: those that can formalize (eg. \texttt{gpt-4o} and \texttt{o3-mini}) and those that cannot formalize (eg. all Llama models and \texttt{gpt-4o-mini}). Among the models that can formalize, we find that the results are robust to lexical perturbations. %This suggests that \textbf{LLM-as-formalizer is robust to lexical perturbation}, and its success is not due to memorization of the domain which is a part of the pretraining data.
%\subsection{Open-sourced Models}
\subsection{What kind of errors?} \label{sec:analysis_errors}

In this section, we discuss the kinds of errors made in PDDL generation. We perform an error analysis on a random 20 sample subset of problems where a plan was not found, or the found plan was not correct. From there, we categorize the errors by syntax errors in either file, semantic errors in the \df, and semantic errors in the \pf. Of the errors in the \df, we determine finer-grained errors such as incorrect or missing actions, preconditions, effects, predicates, and parameters. The error analysis for both Natural \bw and \lgt can be found in Figure~\ref{fig:analysis of errors}. 

\textbf{For the open-source models, the most common error is syntax errors} on \bw and \lgt. For example, models repeatedly use the keyword `preconditions' instead of `precondition' which might suggest a lack of grasp of the PDDL syntax. Gemma models like \texttt{gemma-2-27b} make significantly fewer syntax errors (3 out of 20 on \bw) than Llama models like \texttt{Llama-3.1-70B} (20 out of 20), despite being smaller. Despite the syntax errors, \textbf{open-sourced models still make many semantic errors in the \df and \pf},  which include missing predicates and incorrect effects in the actions. We also observe a significant gap between the number of plans that were found, and the number of found plans that were correct. We find that the most common error made was swapping the parameters in the preconditions of the `stack' action, leading to incorrect plans. 
While making much fewer syntax errors, \textbf{GPT models frequently suffer from semantic errors}. Interestingly, the most common error made for \texttt{gpt-4o} come from the \pf, which is intuitively easier to generate than the \df. A common error in the \pf is incorrect predicates in the initial state and goal state. A common error in the \df is incorrect effects in the actions. For example, in the `unstack' action, the model does not make the next block `clear' when the top block has been placed in the hand. 
For \mbw, there are barely any syntax errors or semantic errors in the \pf but rather the most common errors come from the \df. Since this domain is a result of lexical perturbation, formalizing in PDDL is akin to symbolic information extraction and translation, devoid of much use of commonsense knowledge. Due to the Heavily Templated descriptions, all the predicates would be listed out in the \pd and the model would just need to match them to PDDL syntax in the \pf. While a similar essence, this is more challenging for the \df since the clauses of preconditions and effects are more involved. From Table~\ref{tab:analysis of df errors}, a similar trend between \bw and \mbw also suggests that the LLM-as-formalizer methodology is robust to such perturbation. 

%As the descriptions get more templated, we see that the amount of errors in the \pf decreases and the errors in the \df increases. This is most likely due to the templated \pd having all the needed predicates in the inital and goal states for the model to generate correct PDDL. 

\definecolor{darkblue}{RGB}{35,50,99}
\definecolor{regblue}{RGB}{188,206,225}
\begin{figure}[t!]
\centering
\small
\begin{tikzpicture}
\begin{axis}[
    ybar stacked,
    bar width=15pt,
    width=\columnwidth,
    height=7cm,
    enlarge x limits=0.15,
    symbolic x coords={gemma-2b, gemma-27b, llama-8b, llama-70b, gpt-4o-mini, gpt-4o},
    xtick=data,
    xticklabel style={rotate=20}, % ← rotate 45°
    ymin=0, ymax=40,
    legend style={at={(0.5,-0.18)}, anchor=north, legend columns=-1},
    every node near coord/.append style={font=\small},
]

% Example values for 3 portions per dataset
\addplot+[fill=darkblue,draw=darkblue] plot coordinates {(gemma-2b,22) (gemma-27b,11) (llama-8b,40) (llama-70b,40) (gpt-4o-mini,4) (gpt-4o,7)};
\addplot+[fill=regblue,draw=regblue] plot coordinates {(gemma-2b,18) (gemma-27b,29) (llama-8b,0) (llama-70b,0) (gpt-4o-mini,36) (gpt-4o,33)};

\legend{Has syntax err., No syntax error but has semantic err.}
\end{axis}
\end{tikzpicture}
\caption{Number of syntax errors (solver cannot parse PDDL) and semantic errors (solver outputs either no plan or a wrong plan) in 40 manually annotated, randomly sampled errors in Natural \bw and \lgt. See full results in Table~\ref{tab:analysis of errors}.}
\label{fig:analysis of errors}
\end{figure}

\begin{figure}[t!]
    \centering
    \small
\begin{tikzpicture}
\begin{axis}[
    ybar,                                  % bar chart
    bar width=12pt,                        % width of each bar
    width=\columnwidth,
    height=7cm,
    ymin=0, ymax=40,                       % y‑axis range
    symbolic x coords={Missing Predicate,Missing Action,Wrong Parameters,Wrong Preconditions,Wrong Effects},    % x‑axis categories
    xtick=data,                            % one tick per category
    xticklabel style={rotate=20, anchor=east}, % ← rotate 45°
    enlarge x limits=0.25,                 % extra space left & right
    legend style={
        at={(0.7,-0.18)},                  % place legend below plot
        anchor=north,
        legend columns=-1
    }
]
  % ---------- Group G1 ----------
  \addplot+[ybar,fill=darkblue,draw=darkblue] coordinates {
    (Missing Predicate,39) (Missing Action,6) (Wrong Parameters, 19) (Wrong Preconditions,31) (Wrong Effects, 34)
  };
  % ---------- Group G2 ----------
  \addplot+[ybar,fill=regblue,draw=regblue] coordinates {
    (Missing Predicate,19) (Missing Action,1) (Wrong Parameters,9) (Wrong Preconditions, 17) (Wrong Effects,11)
  };

  \legend{gpt-4o-mini,gpt-4o}
\end{axis}
\end{tikzpicture}
    \caption{Number of fine-grained \df errors in 40 manually annotated, randomly sampled errors in Natural \bw and \lgt. See full results in Table~\ref{tab:analysis of df errors}.}
    \label{fig:enter-label}
\end{figure}

\subsection{The curious case of \texttt{DeepSeek-R1}}
The only exception to our main findings is \texttt{DeepSeek-R1} for all datasets. The model shows strong performance as both a formalizer and planner for \bw, while showing mixed results for \lgt, and following a similar trend as GPT models for \mbw. We can conclude that for simple datasets, using \texttt{DeepSeek-R1} as a formalizer is preferred for some datasets, and when formalizing, the model is robust to lexical perturbations. However, for more complex datasets, using the model as a planner is preferred. This result may be due to how \texttt{DeepSeek-R1} generates its response. The model outputs a reasoning string that thinks through every step of the plan and attempts to check its work by stepping through the plan. After examining several reasoning strings, we notice that when we use the model as a planner, it will think through the plan and when it gets stuck, it will attempt to pattern-match the example output given in the prompt. While the example output for each domain is not found in any of the datasets, we believe this may have caused the higher performance on more complex domains.

\section{Related Work}

\textbf{Planning with LLMs} There has been a large amount of research using LLMs for planning tasks. Some use LLMs for informal planning, also known as script or procedure learning \cite{zhang-etal-2020-analogous,lyu-etal-2021-goal,lal-etal-2024-tailoring}. While modern LLMs can make coherent and plausible informal plans, they are ungrounded and so lack executability and verifiability. Work that uses LLMs for formal planning in grounded environments generally concludes with the inability of such LLMs-as-planners \cite{silver2024generalized,valmeekam2024planbench,stechly2024chain,valmeekam2024llmscantplanlrms}. Follow-up work tackles this shortcoming by using the LLM as a heuristic, not just a planner, such as by proposing candidate plans that are iteratively verified \cite{valmeekam2023planningabilitieslargelanguage,kambhampati2024llms}. While we consider the standard LLM-as-planner as a baseline, our focus is on LLM-as-formalizer, an alternative methodology for the same problem.

\noindent \textbf{LLMs as PDDL formalizer} Here, LLMs do not provide plans but rather generate a PDDL representation of the domain and problem, which is then run through a solver to find the plan. This methodology has proven successful in a number of recent works, where the LLM generates different parts but not all of the PDDL for simplified evaluation. \citet{zuo2024planetarium, zhang-etal-2024-pddlego, liu2023llm+} use the LLM to predict the entire \pf, while \citet{xie2023translating, lyu-etal-2023-faithful} predict just the goal for the \pf. Some predict parts of the \df, such as \citet{zhang-etal-2024-proc2pddl, zhu2024language} which generate the action semantics of the \df and \citet{wong2023learning} who also predicts the predicates from a candidate list. Closest to our work is \citet{guan2023leveraging} which predicts the \df as well as the \pf goal. However, our work of holistically generating PDDL shows that coming up with the initial state in the \pf is non-trivial (Section~\ref{sec:analysis_errors}). Moreover, we vary the level of naturalness of descriptions in addition to the templated ones, which prove to be more challenging and insightful (Section~\ref{sec:analysis_naturalness}).

While the above discussions pertain to LLMs generating PDDL, many work on embodied agents outside the NLP community tackle similar problems with different focus \cite{li2024embodied}.

\noindent \textbf{LLM code generation} Our work hinges on modern LLMs' ability to generate code \cite{chen2021evaluating}. In addition to writing or debugging programs \cite{jiang2024survey}, LLMs are also used to generate formal, interim representations that are not necessarily PDDL for problem solving. For example, \citet{gao2023pal,lyu-etal-2023-faithful,tang2024worldcodermodelbasedllmagent} use the LLM to generate executable Python code for solving symbolic problems. In other work, the generated code may not be executable and is provided to another LLMs to facilitate reasoning \cite{madaan-etal-2022-language,zhang-etal-2023-causal}.

A table comparing a couple of these works can be seen in Table \ref{tab:related_works} in the Appendix (Section \ref{related works comparison}).

\iffalse
\textbf{LLMs and Naturalness}
A key part of our research is varying the naturalness of the language in the problem to see if the model can translate the problem into PDDL using only the information that a human needs. There has been research to learn how much language does a model fully understand when solving problems, but there has not been research on how much the LLM can understand to generate into PDDL. 
\fi
%0.75 page. 

%Possible areas to discuss:
%\begin{enumerate}
   % \item Informal planning
   % \item LLM-as-planner (how it doesn't work)
    %\item LLM-as-formalizer (PDDL, non-PDDL; domain modeling, goal interpretation (can reference Manling's and Zhu's paper's comparison table))
    %\item Program synthesis, code generation and neurosymbolic methods
%\end{enumerate}

\section{Conclusion and Future Work}

Amidst the trend of using LLMs as formalizers, especially to generate PDDL for planning, our comprehensive analysis can be distilled into the following key findings and suggestions for future work.

\noindent \textbf{State-of-the-art LLMs can plan and formalize} at a high level on simple domains like BlocksWorld, and to some degree on more complex domains like Logistics. Moreover, reasoning models trained to produce reasoning tokens before the answer (such as \texttt{o3-mini} and \texttt{DeepSeek-R1}) are superior planners than those who are not (such as \texttt{gpt-4o}). We suggest future work closely study these models (such as \citet{valmeekam2024llmscantplanlrms}) from various angles such as the soundness of their reasoning chain, time and monetary cost, etc. 

\noindent \textbf{Most small open LLMs cannot plan nor formalize} even on simple domains, succumbing to syntax errors. To improve their performance on formalizing, we suggest future efforts on methods such as finetuning, modular prompting, constrained decoding, retrieval by language documentation, wrapper of a high-resource language, and so on to improve the performance of generating not just PDDL, but also low-resource languages in general. 

\noindent \textbf{Formalizing is often superior to planning, but not always}, with positive evidence for models like \texttt{gpt-4o} but negative evidence for models like \texttt{DeepSeek-R1}. Given the rising attention towards the LLM-as-formalizer methodology, we suggest future work validate this claim and extend to other environments such as more complex ones, partially observable ones, etc.

\noindent \textbf{Natural descriptions are challenging} in planning tasks, and \textbf{so is predicting complete PDDL}. We acknowledge that translating a simple natural language instruction to a specific PDDL component might be sufficient in some confined domains and a fruitful endeavor. However, we argue that generating a planning representation as complete as possible given a natural language query as complex and realistic as possible is a prerequisite to open-domain planning.

%We explore the limit of state-of-the-art LLMs to be used as a PDDL formalizer for planning with natural language descriptions of different naturalness levels. We conclude that with zero-shot prompting, only GPT models as formalizer are sufficiently capable in multiple domains. Therefore, it is important to equip open-source models with such ability to democratize the ability of making executable plans. We also show that LLM-as-formalizer is robust to lexical perturbation, demonstrating strong performance in long-tail domains that are underrepresented in pretraining. Our work will inform future efforts of using LLM as a planning formalizer, including experiments on partially-observed environments that require exploration and interaction, more complex environments with a larger action space, and so on.

\section{Limitation}
%While BlocksWorld is one of the most frequently used domains in LLM's formal planning, it is the only domain considered in this work, due to the choice to optimize for the analysis' depth instead of breadth. As future work, we are currently considering complex domains such as Barman in IPC. 

A common and valid criticism for using these simulations or text problems for evaluation is that their settings may be too contrived and removed from reality. Nevertheless, it is likely that LLMs' satisfactory performance on these datasets is a necessary condition to success in real life. 

While we only consider zero-shot prompting, and a few experiments with few-shot and chain-of-thought prompts, without any attempt for prompt tuning, it is possible that the models' performance significantly increases otherwise. Therefore, experimental results in all settings may be underestimated. Moreover, advanced prompting techniques such as self-refine, and voting can all potentially improve model performance. However, the study of those is out of the scope of this work.

While we advocate for the LLM-as-formalizer methodology over LLM-as-planner, the former's success may be dependent on the task. Highly symbolic tasks which can be relatively easily described, like BlocksWorld, are likely to favor LLM-as-formalizer. However, LLM-as-planner might shine in tasks with a more complex action space requiring common-sense knowledge that is easily accessed by pretraining. Furthermore, while we only consider the most straightforward LLM-as-planner prompting method, more involved methods, like \citet{kambhampati2024llms} that combines LLM-as-planner with symbolic validation, will likely lead to a stronger baseline.

Since this work uses only the BlocksWorld, Mystery BlocksWorld, Logistics and Barman domains, it is a small toy example to the usage of LLMs as formalizers and are not representative to problems in the real world, which would be much more challenging. This may pose a risk to users using this code on real world problems.

The datasets we use and we propose are all under the MIT License.

\section*{Acknowledgment}
We thank Peter Clark for providing invaluable input throughout this work and Andrew Zhu for providing technical support on LLM inference.

\bibliography{anthology, custom}

\begin{thebibliography}{40}
\providecommand{\natexlab}[1]{#1}

\bibitem[{Berko(1958)}]{berko1958child}
Jean Berko. 1958.
\newblock The child's learning of english morphology.
\newblock \emph{Word}, 14(2-3):150--177.

\bibitem[{Cassano et~al.(2022)Cassano, Gouwar, Nguyen, Nguyen, Phipps-Costin, Pinckney, Yee, Zi, Anderson, Feldman et~al.}]{cassano2022multipl}
Federico Cassano, John Gouwar, Daniel Nguyen, Sydney Nguyen, Luna Phipps-Costin, Donald Pinckney, Ming-Ho Yee, Yangtian Zi, Carolyn~Jane Anderson, Molly~Q Feldman, et~al. 2022.
\newblock Multipl-e: A scalable and extensible approach to benchmarking neural code generation.
\newblock \emph{arXiv preprint arXiv:2208.08227}.

\bibitem[{Chen et~al.(2021)Chen, Tworek, Jun, Yuan, Pinto, Kaplan, Edwards, Burda, Joseph, Brockman et~al.}]{chen2021evaluating}
Mark Chen, Jerry Tworek, Heewoo Jun, Qiming Yuan, Henrique Ponde De~Oliveira Pinto, Jared Kaplan, Harri Edwards, Yuri Burda, Nicholas Joseph, Greg Brockman, et~al. 2021.
\newblock Evaluating large language models trained on code.
\newblock \emph{arXiv preprint arXiv:2107.03374}.

\bibitem[{Dubey et~al.(2024)Dubey, Jauhri, Pandey, Kadian, Al-Dahle, Letman, Mathur, Schelten, Yang, Fan et~al.}]{dubey2024llama}
Abhimanyu Dubey, Abhinav Jauhri, Abhinav Pandey, Abhishek Kadian, Ahmad Al-Dahle, Aiesha Letman, Akhil Mathur, Alan Schelten, Amy Yang, Angela Fan, et~al. 2024.
\newblock The llama 3 herd of models.
\newblock \emph{arXiv preprint arXiv:2407.21783}.

\bibitem[{Gao et~al.(2023)Gao, Madaan, Zhou, Alon, Liu, Yang, Callan, and Neubig}]{gao2023pal}
Luyu Gao, Aman Madaan, Shuyan Zhou, Uri Alon, Pengfei Liu, Yiming Yang, Jamie Callan, and Graham Neubig. 2023.
\newblock Pal: Program-aided language models.
\newblock In \emph{International Conference on Machine Learning}, pages 10764--10799. PMLR.

\bibitem[{Guan et~al.(2023)Guan, Valmeekam, Sreedharan, and Kambhampati}]{guan2023leveraging}
Lin Guan, Karthik Valmeekam, Sarath Sreedharan, and Subbarao Kambhampati. 2023.
\newblock Leveraging pre-trained large language models to construct and utilize world models for model-based task planning.
\newblock \emph{Advances in Neural Information Processing Systems}, 36:79081--79094.

\bibitem[{Guo et~al.(2025)Guo, Yang, Zhang, Song, Zhang, Xu, Zhu, Ma, Wang, Bi et~al.}]{guo2025deepseek}
Daya Guo, Dejian Yang, Haowei Zhang, Junxiao Song, Ruoyu Zhang, Runxin Xu, Qihao Zhu, Shirong Ma, Peiyi Wang, Xiao Bi, et~al. 2025.
\newblock Deepseek-r1: Incentivizing reasoning capability in llms via reinforcement learning.
\newblock \emph{arXiv preprint arXiv:2501.12948}.

\bibitem[{Howey et~al.(2004)Howey, Long, and Fox}]{1374201}
R.~Howey, D.~Long, and M.~Fox. 2004.
\newblock \href {https://doi.org/10.1109/ICTAI.2004.120} {Val: automatic plan validation, continuous effects and mixed initiative planning using pddl}.
\newblock In \emph{16th IEEE International Conference on Tools with Artificial Intelligence}, pages 294--301.

\bibitem[{IPC(1998)}]{ipc}
IPC. 1998.
\newblock International planning competition.
\newblock \url{https://www.icaps-conference.org/competitions}.

\bibitem[{IPC(2000)}]{ipc-2000}
IPC. 2000.
\newblock International planning competition.
\newblock \url{https://www.icaps-conference.org/competitions}.

\bibitem[{Jiang et~al.(2024)Jiang, Wang, Shen, Kim, and Kim}]{jiang2024survey}
Juyong Jiang, Fan Wang, Jiasi Shen, Sungju Kim, and Sunghun Kim. 2024.
\newblock A survey on large language models for code generation.
\newblock \emph{arXiv preprint arXiv:2406.00515}.

\bibitem[{Kambhampati et~al.(2024)Kambhampati, Valmeekam, Guan, Verma, Stechly, Bhambri, Saldyt, and Murthy}]{kambhampati2024llms}
Subbarao Kambhampati, Karthik Valmeekam, Lin Guan, Mudit Verma, Kaya Stechly, Siddhant Bhambri, Lucas Saldyt, and Anil Murthy. 2024.
\newblock Llms can't plan, but can help planning in llm-modulo frameworks.
\newblock \emph{arXiv preprint arXiv:2402.01817}.

\bibitem[{Lal et~al.(2024)Lal, Zhang, Brahman, Majumder, Clark, and Tandon}]{lal-etal-2024-tailoring}
Yash~Kumar Lal, Li~Zhang, Faeze Brahman, Bodhisattwa~Prasad Majumder, Peter Clark, and Niket Tandon. 2024.
\newblock \href {https://doi.org/10.18653/v1/2024.findings-acl.921} {Tailoring with targeted precision: Edit-based agents for open-domain procedure customization}.
\newblock In \emph{Findings of the Association for Computational Linguistics: ACL 2024}, pages 15597--15611, Bangkok, Thailand. Association for Computational Linguistics.

\bibitem[{Li et~al.(2024)Li, Zhao, Wang, Wang, Zhou, Srivastava, Gokmen, Lee, Li, Zhang et~al.}]{li2024embodied}
Manling Li, Shiyu Zhao, Qineng Wang, Kangrui Wang, Yu~Zhou, Sanjana Srivastava, Cem Gokmen, Tony Lee, Li~Erran Li, Ruohan Zhang, et~al. 2024.
\newblock Embodied agent interface: Benchmarking llms for embodied decision making.
\newblock \emph{arXiv preprint arXiv:2410.07166}.

\bibitem[{Liu et~al.(2023{\natexlab{a}})Liu, Jiang, Zhang, Liu, Zhang, Biswas, and Stone}]{liu2023llm+}
Bo~Liu, Yuqian Jiang, Xiaohan Zhang, Qiang Liu, Shiqi Zhang, Joydeep Biswas, and Peter Stone. 2023{\natexlab{a}}.
\newblock Llm+ p: Empowering large language models with optimal planning proficiency.
\newblock \emph{arXiv preprint arXiv:2304.11477}.

\bibitem[{Liu et~al.(2023{\natexlab{b}})Liu, Ning, Teng, Liu, Zhou, and Zhang}]{liu2023evaluating}
Hanmeng Liu, Ruoxi Ning, Zhiyang Teng, Jian Liu, Qiji Zhou, and Yue Zhang. 2023{\natexlab{b}}.
\newblock Evaluating the logical reasoning ability of chatgpt and gpt-4.
\newblock \emph{arXiv preprint arXiv:2304.03439}.

\bibitem[{Lyu et~al.(2023)Lyu, Havaldar, Stein, Zhang, Rao, Wong, Apidianaki, and Callison-Burch}]{lyu-etal-2023-faithful}
Qing Lyu, Shreya Havaldar, Adam Stein, Li~Zhang, Delip Rao, Eric Wong, Marianna Apidianaki, and Chris Callison-Burch. 2023.
\newblock \href {https://doi.org/10.18653/v1/2023.ijcnlp-main.20} {Faithful chain-of-thought reasoning}.
\newblock In \emph{Proceedings of the 13th International Joint Conference on Natural Language Processing and the 3rd Conference of the Asia-Pacific Chapter of the Association for Computational Linguistics (Volume 1: Long Papers)}, pages 305--329, Nusa Dua, Bali. Association for Computational Linguistics.

\bibitem[{Lyu et~al.(2021)Lyu, Zhang, and Callison-Burch}]{lyu-etal-2021-goal}
Qing Lyu, Li~Zhang, and Chris Callison-Burch. 2021.
\newblock \href {https://doi.org/10.18653/v1/2021.inlg-1.19} {Goal-oriented script construction}.
\newblock In \emph{Proceedings of the 14th International Conference on Natural Language Generation}, pages 184--200, Aberdeen, Scotland, UK. Association for Computational Linguistics.

\bibitem[{Madaan et~al.(2022)Madaan, Zhou, Alon, Yang, and Neubig}]{madaan-etal-2022-language}
Aman Madaan, Shuyan Zhou, Uri Alon, Yiming Yang, and Graham Neubig. 2022.
\newblock \href {https://doi.org/10.18653/v1/2022.emnlp-main.90} {Language models of code are few-shot commonsense learners}.
\newblock In \emph{Proceedings of the 2022 Conference on Empirical Methods in Natural Language Processing}, pages 1384--1403, Abu Dhabi, United Arab Emirates. Association for Computational Linguistics.

\bibitem[{Muise(2016)}]{muise-icaps16demo-pd}
Christian Muise. 2016.
\newblock {Planning.Domains}.
\newblock In \emph{The 26th International Conference on Automated Planning and Scheduling - Demonstrations}.

\bibitem[{Pan et~al.(2023)Pan, Albalak, Wang, and Wang}]{pan-etal-2023-logic}
Liangming Pan, Alon Albalak, Xinyi Wang, and William Wang. 2023.
\newblock \href {https://doi.org/10.18653/v1/2023.findings-emnlp.248} {Logic-{LM}: Empowering large language models with symbolic solvers for faithful logical reasoning}.
\newblock In \emph{Findings of the Association for Computational Linguistics: EMNLP 2023}, pages 3806--3824, Singapore. Association for Computational Linguistics.

\bibitem[{Seipp et~al.(2022)Seipp, Torralba, and Hoffmann}]{seipp-et-al-zenodo2022}
Jendrik Seipp, {\'A}lvaro Torralba, and J{\"o}rg Hoffmann. 2022.
\newblock {PDDL} generators.
\newblock \url{https://doi.org/10.5281/zenodo.6382173}.

\bibitem[{Silver et~al.(2024)Silver, Dan, Srinivas, Tenenbaum, Kaelbling, and Katz}]{silver2024generalized}
Tom Silver, Soham Dan, Kavitha Srinivas, Joshua~B Tenenbaum, Leslie Kaelbling, and Michael Katz. 2024.
\newblock Generalized planning in pddl domains with pretrained large language models.
\newblock In \emph{Proceedings of the AAAI Conference on Artificial Intelligence}, volume~38, pages 20256--20264.

\bibitem[{Stechly et~al.(2024)Stechly, Valmeekam, and Kambhampati}]{stechly2024chain}
Kaya Stechly, Karthik Valmeekam, and Subbarao Kambhampati. 2024.
\newblock Chain of thoughtlessness: An analysis of cot in planning.
\newblock \emph{arXiv preprint arXiv:2405.04776}.

\bibitem[{Tang et~al.(2024)Tang, Key, and Ellis}]{tang2024worldcodermodelbasedllmagent}
Hao Tang, Darren Key, and Kevin Ellis. 2024.
\newblock \href {https://arxiv.org/abs/2402.12275} {Worldcoder, a model-based llm agent: Building world models by writing code and interacting with the environment}.
\newblock \emph{Preprint}, arXiv:2402.12275.

\bibitem[{Team et~al.(2024)Team, Riviere, Pathak, Sessa, Hardin, Bhupatiraju, Hussenot, Mesnard, Shahriari, Ram{\'e} et~al.}]{team2024gemma}
Gemma Team, Morgane Riviere, Shreya Pathak, Pier~Giuseppe Sessa, Cassidy Hardin, Surya Bhupatiraju, L{\'e}onard Hussenot, Thomas Mesnard, Bobak Shahriari, Alexandre Ram{\'e}, et~al. 2024.
\newblock Gemma 2: Improving open language models at a practical size.
\newblock \emph{arXiv preprint arXiv:2408.00118}.

\bibitem[{Valmeekam et~al.(2024{\natexlab{a}})Valmeekam, Marquez, Olmo, Sreedharan, and Kambhampati}]{valmeekam2024planbench}
Karthik Valmeekam, Matthew Marquez, Alberto Olmo, Sarath Sreedharan, and Subbarao Kambhampati. 2024{\natexlab{a}}.
\newblock Planbench: An extensible benchmark for evaluating large language models on planning and reasoning about change.
\newblock \emph{Advances in Neural Information Processing Systems}, 36.

\bibitem[{Valmeekam et~al.(2023)Valmeekam, Sreedharan, Marquez, Olmo, and Kambhampati}]{valmeekam2023planningabilitieslargelanguage}
Karthik Valmeekam, Sarath Sreedharan, Matthew Marquez, Alberto Olmo, and Subbarao Kambhampati. 2023.
\newblock \href {https://arxiv.org/abs/2302.06706} {On the planning abilities of large language models (a critical investigation with a proposed benchmark)}.
\newblock \emph{Preprint}, arXiv:2302.06706.

\bibitem[{Valmeekam et~al.(2024{\natexlab{b}})Valmeekam, Stechly, and Kambhampati}]{valmeekam2024llmscantplanlrms}
Karthik Valmeekam, Kaya Stechly, and Subbarao Kambhampati. 2024{\natexlab{b}}.
\newblock \href {https://arxiv.org/abs/2409.13373} {Llms still can't plan; can lrms? a preliminary evaluation of openai's o1 on planbench}.
\newblock \emph{Preprint}, arXiv:2409.13373.

\bibitem[{Weld(1999)}]{weld1999recent}
Daniel~S Weld. 1999.
\newblock Recent advances in ai planning.
\newblock \emph{AI magazine}, 20(2):93--93.

\bibitem[{Wong et~al.(2023)Wong, Mao, Sharma, Siegel, Feng, Korneev, Tenenbaum, and Andreas}]{wong2023learning}
Lionel Wong, Jiayuan Mao, Pratyusha Sharma, Zachary~S Siegel, Jiahai Feng, Noa Korneev, Joshua~B Tenenbaum, and Jacob Andreas. 2023.
\newblock Learning adaptive planning representations with natural language guidance.
\newblock \emph{arXiv preprint arXiv:2312.08566}.

\bibitem[{Xie et~al.(2023)Xie, Yu, Zhu, Bai, Gong, and Soh}]{xie2023translating}
Yaqi Xie, Chen Yu, Tongyao Zhu, Jinbin Bai, Ze~Gong, and Harold Soh. 2023.
\newblock Translating natural language to planning goals with large-language models.
\newblock \emph{arXiv preprint arXiv:2302.05128}.

\bibitem[{Zhang et~al.(2020)Zhang, Chen, Wang, Song, and Roth}]{zhang-etal-2020-analogous}
Hongming Zhang, Muhao Chen, Haoyu Wang, Yangqiu Song, and Dan Roth. 2020.
\newblock \href {https://doi.org/10.18653/v1/2020.emnlp-main.119} {Analogous process structure induction for sub-event sequence prediction}.
\newblock In \emph{Proceedings of the 2020 Conference on Empirical Methods in Natural Language Processing (EMNLP)}, pages 1541--1550, Online. Association for Computational Linguistics.

\bibitem[{Zhang et~al.(2024{\natexlab{a}})Zhang, Jansen, Zhang, Clark, Callison-Burch, and Tandon}]{zhang-etal-2024-pddlego}
Li~Zhang, Peter Jansen, Tianyi Zhang, Peter Clark, Chris Callison-Burch, and Niket Tandon. 2024{\natexlab{a}}.
\newblock \href {https://doi.org/10.18653/v1/2024.starsem-1.17} {{PDDLEGO}: Iterative planning in textual environments}.
\newblock In \emph{Proceedings of the 13th Joint Conference on Lexical and Computational Semantics (*SEM 2024)}, pages 212--221, Mexico City, Mexico. Association for Computational Linguistics.

\bibitem[{Zhang et~al.(2024{\natexlab{b}})Zhang, Xu, Kommula, Callison-Burch, and Tandon}]{zhang-etal-2024-openpi2}
Li~Zhang, Hainiu Xu, Abhinav Kommula, Chris Callison-Burch, and Niket Tandon. 2024{\natexlab{b}}.
\newblock \href {https://aclanthology.org/2024.eacl-long.10} {{O}pen{PI}2.0: An improved dataset for entity tracking in texts}.
\newblock In \emph{Proceedings of the 18th Conference of the European Chapter of the Association for Computational Linguistics (Volume 1: Long Papers)}, pages 166--178, St. Julian{'}s, Malta. Association for Computational Linguistics.

\bibitem[{Zhang et~al.(2023)Zhang, Xu, Yang, Zhou, You, Arora, and Callison-Burch}]{zhang-etal-2023-causal}
Li~Zhang, Hainiu Xu, Yue Yang, Shuyan Zhou, Weiqiu You, Manni Arora, and Chris Callison-Burch. 2023.
\newblock \href {https://doi.org/10.18653/v1/2023.findings-eacl.31} {Causal reasoning of entities and events in procedural texts}.
\newblock In \emph{Findings of the Association for Computational Linguistics: EACL 2023}, pages 415--431, Dubrovnik, Croatia. Association for Computational Linguistics.

\bibitem[{Zhang et~al.(2024{\natexlab{c}})Zhang, Zhang, Hou, Wang, Gu, Clark, Callison-Burch, and Tandon}]{zhang-etal-2024-proc2pddl}
Tianyi Zhang, Li~Zhang, Zhaoyi Hou, Ziyu Wang, Yuling Gu, Peter Clark, Chris Callison-Burch, and Niket Tandon. 2024{\natexlab{c}}.
\newblock \href {https://aclanthology.org/2024.nlrse-1.2} {{PROC}2{PDDL}: Open-domain planning representations from texts}.
\newblock In \emph{Proceedings of the 2nd Workshop on Natural Language Reasoning and Structured Explanations (@ACL 2024)}, pages 13--24, Bangkok, Thailand. Association for Computational Linguistics.

\bibitem[{Zhu et~al.(2023)Zhu, Dugan, Hwang, and Callison-Burch}]{zhu-etal-2023-kani}
Andrew Zhu, Liam Dugan, Alyssa Hwang, and Chris Callison-Burch. 2023.
\newblock \href {https://doi.org/10.18653/v1/2023.nlposs-1.8} {Kani: A lightweight and highly hackable framework for building language model applications}.
\newblock In \emph{Proceedings of the 3rd Workshop for Natural Language Processing Open Source Software (NLP-OSS 2023)}, pages 65--77, Singapore. Association for Computational Linguistics.

\bibitem[{Zhu et~al.(2024)Zhu, Singh, Jia, and Thomason}]{zhu2024language}
Wang Zhu, Ishika Singh, Robin Jia, and Jesse Thomason. 2024.
\newblock Language models can infer action semantics for classical planners from environment feedback.
\newblock \emph{arXiv preprint arXiv:2406.02791}.

\bibitem[{Zuo et~al.(2024)Zuo, Velez, Li, Littman, and Bach}]{zuo2024planetarium}
Max Zuo, Francisco~Piedrahita Velez, Xiaochen Li, Michael~L Littman, and Stephen~H Bach. 2024.
\newblock Planetarium: A rigorous benchmark for translating text to structured planning languages.
\newblock \emph{arXiv preprint arXiv:2407.03321}.

\end{thebibliography}

\appendix

\section{Data Examples}
\label{sec:data_examples}

As discussed above, each dataset comes with ground-truth PDDL describing domains (\df) and problems (\pf). To maximize flexibility when performing analysis, we construct problem instances ourselves for some datasets, so that we can measure complexity with metrics like the number of blocks in BlocksWorld, for which we ensure a uniform distribution to avoid biases. Instances for BlocksWorld were randomly generated by varying the number of blocks and number of stacks in the initial and goal states from 2 to 15. Instances for Mystery BlocksWorld were randomly sampled from \cite{valmeekam2024planbench}. Instances of Logistics were taken directly from \cite{ipc} and \cite{ipc-2000}. Instances of Barman were generated by using \cite{seipp-et-al-zenodo2022} and varying the number of shot-glasses, ingredients and cocktails from 1 to 9. Below are the example PDDL and descriptions for \bw and \mbw. Due to the length of the examples, the PDDL and natural language descriptions for \lgt and \bm can be found at \url{https://github.com/CassieHuang22/llm-as-pddl-formalizer/tree/main/examples}.

\subsection{\bw PDDL}
Listings \ref{lst:blocksworld_df} and \ref{lst:blocksworld_pf} display examples of the ground-truth \df and \pf for \bw.

The \df contains all four actions (pickup, putdown, stack and unstack) and their preconditions and post-conditions, as well as predicates needed for the domain.

The \pf contains the objects, initial state and goal state for the problem.

\subsection{\bw \dd and \pd}
Listings \ref{lst:heavily_templated_blocksworld_dd} and \ref{lst:heavily_templated_blocksworld_pd} display example \dd and \pd for the Heavily Templated setting in the \bw dataset. In this setting, all preconditions and post-conditions are written out explicitly and sound similar to PDDL. The \pd is similar, in that it lists all the predicates needed for to solve the task.

Listings \ref{lst:moderately_templated_blocksworld_dd} and \ref{lst:moderately_templated_blocksworld_pd} display example \dd and \pd for Moderately Templated data in the \bw dataset. In this setting, the \dd and \pd are much more natural than the Heavily Templated data, but all predicates are still listed.

Finally Listings \ref{lst:natural_blocksworld_dd} and \ref{lst:natural_blocksworld_pd} display example \dd and \pd for Natural \bw data. For the Natural data, we can see that the \dd and \pd give all necessary information to complete the task, but does not sound like PDDL, and does not describe all predicates needed to perform the task.

We can see that the descriptions have all the same components as the \df and \pf in PDDL, but written in different levels of naturalness.

\subsection{\mbw PDDL}
Listings \ref{lst:mystery_blocksworld_df} and \ref{lst:mystery_blocksworld_pf} display examples of the ground-truth \df and \pf for \mbw.

\subsection{\mbw \dd and \pd}
Listings \ref{lst:mystery_blocksworld_dd} and \ref{lst:mystery_blocksworld_pd} are example \dd and \pd of the Heavily Templated \mbw. Text written in \textcolor{OliveGreen}{green} demonstrates natural sounding text while text written in \textcolor{red}{red} demonstrates text that sounds the most like PDDL.

\section{Prompts}
\label{sec:prompts}

%\hc{We already have the prompt for LLM-as-planner, but we still need the prompt for LLM-as-planner.}
Listings \ref{lst:blocksworld_planner}, \ref{lst:mystery_blocksworld_planner}, \ref{lst:logistics_planner} and \ref{lst:barman_planner} displays the prompts for all domains given to all models. Whenever possible, we asked the model to return the output in a JSON object for easier parsing.

\section{Experimental Setup Details} \label{sec:solver info}
\subsection{Planner}
We use a \texttt{dual-bfws-ffparser} planner to find plans from the generated PDDL. This is a best first width search planner used with a fast forward planner.

\begin{table*}[!t]
    \centering
    \resizebox{\textwidth}{!}{\begin{tabular}{llll}
    \toprule
         & Environment & LLM predicts? & Natural Descriptions? \\ \midrule
         \citet{zuo2024planetarium} & fully-observed & \pf & Y \\
         \citet{zhang-etal-2024-pddlego} & partially-observed & \pf & N \\
         \citet{liu2023llm+} & fully-observed & \pf & N \\
         \citet{xie2023translating} & fully-observed \& partially-observed & \pf goal & N \\
         \citet{lyu-etal-2023-faithful} & fully-observed & \pf goal & N\\
         \citet{zhang-etal-2024-proc2pddl} & procedural texts & \df action semantics & N \\
         \citet{wong2023learning} & partially-observed & \df & N \\
         \citet{guan2023leveraging} & fully-observed & \df \& \pf goal & N \\
         \citet{zhu2024language} & fully-observed & \df action semantics & N \\ \midrule
         \citet{tang2024worldcodermodelbasedllmagent} & partially-observed & Python & N/A \\
         \midrule
         \citet{silver2024generalized} & fully-observed & plan &  N \\
         \citet{valmeekam2024planbench} & fully-observed & plan & N \\
         \citet{stechly2024chain} & fully-observed & plan & N \\ \midrule
         This work & fully-observed & \df \& \pf & Y \\ \bottomrule
    \end{tabular}}
    \caption{Comparison with related work.}
    \label{tab:related_works}
\end{table*}

\subsection{VAL}
The VAL library takes in a ground-truth PDDL \df, \pf and a plan and tries to execute the plan in the environment by checking whether each action in the found plan can be executed based on the preconditions written in the ground-truth files. If the plan is executable, VAL then checks whether the final state after executing the plan matches the goal state found in the ground-truth \pf. If either the plan is not executable or the final state does not match the goal state, VAL will return an error, therefore the plan found by the planner is not correct. 

\subsection{Evaluation Metrics} We use solvability and correctness to evaluate the results from the LLMs. Solvability was calculated by running the planner and counting how many plans were returned. If a plan was not returned due to not finding a plan or an error occurred, it did not count towards solvability. Correctness was calculated by counting how many plans returned with no errors after running VAL, the plan was counted if VAL was able to execute the plan and there were no errors, and the final state is equal to the goal state in ground-truth problem file.

\section{Detailed Results} \label{app:detailed_results}
Beyond the visualizations above, we show the detailed results of all models on all simulations of all naturalness levels. 

\subsection{Results for \bw}

\label{sec:bw_results}

Table \ref{tab:BW-100} displays results for all results for \bw.

\subsection{Results for \mbw}
Table \ref{tab:Heavily Templated MBW-111} displays the results for Heavily Templated \mbw.

\subsection{Results for \lgt}

Table \ref{tab:logistics} displays results for all naturalness settings for \lgt.

\subsection{Results for \bm}

\begin{figure}
    \centering
    \includegraphics[width=\columnwidth]{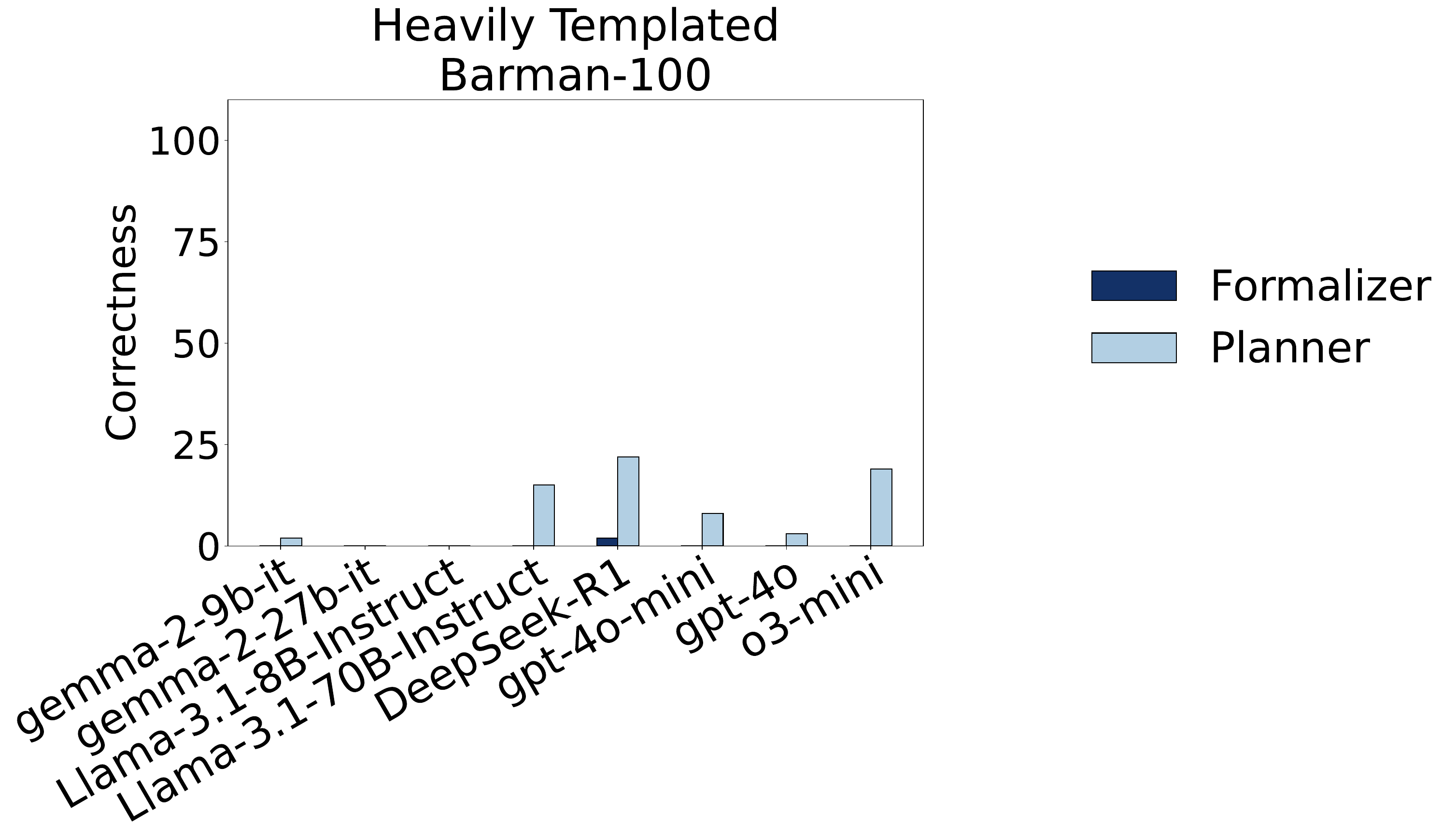}
    \caption{Performance for \bm.}
    \label{fig:results_bm}
\end{figure}
Table \ref{tab:barman} displays results for \bm.

\subsection{Few-Shot Prompting Results} \label{few-shot}
As mentioned in Section \ref{models}, we cannot provide the entire \df and \pf in the prompt for few-shot prompting as we are asking the LLM to predict the entire \df and \pf. Therefore, we performed experiments where we provide a template for the \df and \pf where portions of the PDDL are stubbed out. We performed experiments on Heavily Templated and Natural \bw on a small subset of models. From Tables \ref{tab:BW-100} and \ref{tab:bw_few-shot}, we can see that results are mixed, with performance improving for some models like \texttt{Llama-3.1-8B-Instruct} and worse performance for others like \texttt{gpt-4o} on the Natural \bw dataset. This suggests few-shot prompting does not improve performance for our experiments.

\subsection{Chain-of-Thought Prompting Results} \label{chain-of-thought}
For chain-of-thought prompting, we ask the model to first write a domain description , then write the \df, then write a problem description then write the \pf. These experiments were performed on Heavily Templated and Natural \bw on a small subset of models. From Tables \ref{tab:BW-100} and \ref{tab:bw_cot}, we can see that results are mixed, with performance improvement on some models like \texttt{gpt-3.5-turbo} and performance decline on other models like \texttt{gpt-4o} on the Natural \bw dataset. This suggests that chain-of-thought prompting does not improve performance for our experiments.

\section{Detailed Error Analysis}
\subsection{Syntax and Semantic Errors}
Table~\ref{tab:analysis of errors} displays syntax and semantic errors of models on Natural \bw and \lgt. We do not analyze \bm which is too challenging for models. 

Table~\ref{tab:analysis of df errors} displays fine-grained \df errors of models on Natural \bw and \lgt.

\subsection{Sample Model Output}

%\hc{Any output you want to discuss in Error Analysis can come here.}
We display an example of the \df and \pf that \texttt{Llama-3.1-8B-Instruct} generated. We can see that there are syntax errors, as well as semantic errors in the \df and \pf. We give the model the \dd and \pd in Listing \ref{lst:llama_dd_pd} as input, which returned the \df and \pf shown in Listings \ref{lst:llama_df} and \ref{lst:llama_pf}. Text in \textcolor{red}{red} displays errors outputted from the model. We can see that in the \df there are syntax errors (incorrect keyword ``preconditions'') as well as semantic errors (incorrect predicates in preconditions and effects). For the output \pf there are semantic errors, such as incorrect preconditions (a block cannot be clear and have another block on top of it) in the init section.

\section{Related Works Comparison} \label{related works comparison}

Table~\ref{tab:related_works} compares works related to this paper. We can see that other works as the LLM to predict either the plan, parts of PDDL files and other languages. We can also see that other works have mostly templated natural language descriptions, while this work uses both templated and natural descriptions.

\section{Methodology Comparison}
Both methodologies used in this paper incorporate LLMs into planning. There are pros and cons to each methodology. When using LLM-as-planner, we have a lightweight solution that returns results very quickly. However, due to the lack of reasoning skills in LLMs, they often struggle to create formal plans. Meanwhile using LLM-as-formalizer provides better executability and interpretability, though it uses a solver, which may results in getting results slower. We believe that for performance reasons that LLM-as-formalizer is the superior methodology.

%\noindent\df:

\lstdefinestyle{mystyle}{
    basicstyle=\ttfamily\small,
    breaklines=true,
    frame=single,
    backgroundcolor=\color{gray!5},
    keywordstyle=\color{black},
    commentstyle=\color{gray},
    stringstyle=\color{red},
    showstringspaces=false,
    tabsize=2,
    rulecolor=\color{black},
    framesep=6pt, 
    columns=fullflexible,
}
\lstset{style=mystyle,escapeinside={<@}{@>},breakatwhitespace=false}

\onecolumn
\begin{lstlisting}[language=lisp, caption={Domain File for BlocksWorld-100}, label={lst:blocksworld_df}]
(define (domain blocksworld) 
(:predicates (clear ?x) 
             (on-table ?x)  
             (arm-empty)    
             (holding ?x)   
             (on ?x ?y))    

(:action pickup 
  :parameters (?ob) 
  :precondition (and (clear ?ob) (on-table ?ob) (arm-empty))    
  :effect (and (holding ?ob) (not (clear ?ob)) (not (on-table ?ob)) 
               (not (arm-empty))))  

(:action putdown    
  :parameters  (?ob)    
  :precondition (holding ?ob)   
  :effect (and (clear ?ob) (arm-empty) (on-table ?ob)   
               (not (holding ?ob))))    

(:action stack  
  :parameters  (?ob ?underob)   
  :precondition (and (clear ?underob) (holding ?ob))    
  :effect (and (arm-empty) (clear ?ob) (on ?ob ?underob)    
               (not (clear ?underob)) (not (holding ?ob)))) 

(:action unstack    
  :parameters  (?ob ?underob)   
  :precondition (and (on ?ob ?underob) (clear ?ob) (arm-empty)) 
  :effect (and (holding ?ob) (clear ?underob)   
               (not (on ?ob ?underob)) (not (clear ?ob)) (not (arm-empty)))))   
\end{lstlisting}
\begin{lstlisting}[language=lisp, caption={Problem File for BlocksWorld-100}, label={lst:blocksworld_pf}]
(define (problem blocksworld-p99)   
  (:domain blocksworld) 
  (:objects red blue green yellow ) 
  (:init    
    (on-table red)  
    (on blue red)   
    (clear blue)    
    (on-table green)    
    (on yellow green)   
    (clear yellow)  
    (arm-empty) 
  ) 
  (:goal (and   
    (on-table red)  
    (on green red)  
    (on yellow green)   
    (on blue yellow)    
  ))    
)   
\end{lstlisting}

\begin{lstlisting}[language=lisp, caption={Domain Description for Heavily Templated BlocksWorld-100}, label={lst:heavily_templated_blocksworld_dd}]
I am playing with a set of blocks. Here are the actions I can do

   Pickup block
   Unstack block from another block
   Putdown block
   Stack block on another block

I have the following restrictions on my actions:
    To perform Pickup action, the following facts need to be true: clear block, block on table, arm-empty.
    Once Pickup action is performed the following facts will be true: holding block.
    Once Pickup action is performed the following facts will be false: clear block, block on table, arm-empty.
    To perform Putdown action, the following facts need to be true: holding block.
    Once Putdown action is performed the following facts will be true: clear block, block on table, arm-empty.    
    Once Putdown action is performed the following facts will be false: holding block.
    To perform Stack action, the following needs to be true: clear block2, holding block1.
    Once Stack action is performed the following will be true: arm-empty, clear block1, block1 on block2.
    Once Stack action is performed the following will be false: clear block2, holding block1.
    To perform Unstack action, the following needs to be true: block1 on block2, clear block1, arm-empty.
    Once Unstack action is performed the following will be true: holding block1, clear block2.
    Once Unstack action is performed the following will be false:, block1 on block2, clear block1, arm-empty.
    \end{lstlisting}

    \begin{lstlisting}[language=lisp, caption={Problem Description for Heavily Templated BlocksWorld-100}, label={lst:heavily_templated_blocksworld_pd}]
As initial conditions I have that, the blue block is clear, the yellow block is clear, arm-empty, the blue block is on top of the red block, the yellow block is on top of the green block, the red block is on the table, and the green block is on the table.
My goal is to have that the blue block is on top of the yellow block, the green block is on top of the red block, the yellow block is on top of the green block, and the red block is on the table.
    \end{lstlisting}

    \begin{lstlisting}[language=lisp, caption={Domain Description for Moderately Templated BlocksWorld-100}, label={lst:moderately_templated_blocksworld_dd}]
I am playing with a set of blocks where I need to arrange the blocks into stacks. Here are the actions I can do

   Pick up a block
   Unstack a block from on top of another block
   Put down a block
   Stack a block on top of another block
   
   I have the following restrictions on my actions:
   I can only pick up or unstack one block at a time.
   I can only pick up or unstack a block if my hand is empty.
   I can only pick up a block if the block is on the table and the block is clear. A block is clear if the block has no other blocks on top of it and if the block is not picked up.
   I can only unstack a block from on top of another block if the block I am unstacking was really on top of the other block.
   I can only unstack a block from on top of another block if the block I am unstacking is clear.
   Once I pick up or unstack a block, I am holding the block.
   I can only put down a block that I am holding.
   I can only stack a block on top of another block if I am holding the block being stacked.
   I can only stack a block on top of another block if the block onto which I am stacking the block is clear.
   Once I put down or stack a block, my hand becomes empty.
   Once you stack a block on top of a second block, the second block is no longer clear.
    \end{lstlisting}

    \begin{lstlisting}[language=lisp, caption={Problem Description for Moderately Templated BlocksWorld-100}, label={lst:moderately_templated_blocksworld_pd}]
As initial conditions I have that, the blue block is clear, the yellow block is clear, the hand is empty, the blue block is on top of the red block, the yellow block is on top of the green block, the red block is on the table, and the green block is on the table.
My goal is to have that the blue block is on top of the yellow block, the green block is on top of the red block, the yellow block is on top of the green block, and the red block is on the table.
    \end{lstlisting}

    \begin{lstlisting}[language=lisp, caption={Domain Description for Natural BlocksWorld-100}, label={lst:natural_blocksworld_dd}]
The Blocksworld game involves a set of blocks of different colors, which can be stacked on top of each other or placed on the table. The objective is to move the blocks from an initial configuration to a goal configuration using a series of legal moves. Legal moves in Blocksworld include: picking up a block from the table or from the top of another block, stacking a block onto the table, or stacking a block onto another block. 
    \end{lstlisting}

    \begin{lstlisting}[language=lisp, caption={Problem Description for Natural BlocksWorld-100}, label={lst:natural_blocksworld_pd}]
In this particular game, there are 4 blocks: a red block, a blue block, a green block, and a yellow block. At the start, the red block is on the table, the blue block is on top of the red block, the green block is on the table, and the yellow block is on top of the green block. The goal is to have the red block on the table, the green block on top of the red block, the yellow block on top of the green block, and the blue block on top of the yellow block.
    \end{lstlisting}

    \begin{lstlisting}[language=lisp, caption={Domain File for Mystery\_BlocksWorld-100}, label={lst:mystery_blocksworld_df}]
(define (domain mystery_blocksworld)
(:predicates (province ?x)
             (planet ?x)
             (harmony)
             (pain ?x)
             (craves ?x ?y))

(:action attack
  :parameters (?ob)
  :precondition (and (province ?ob) (planet ?ob) (harmony))
  :effect (and (pain ?ob) (not (province ?ob)) (not (planet ?ob))
               (not (harmony))))

(:action succumb
  :parameters  (?ob)
  :precondition (pain ?ob)
  :effect (and (province ?ob) (harmony) (planet ?ob)
               (not (pain ?ob))))

(:action overcome
  :parameters  (?ob ?underob)
  :precondition (and (province ?underob) (pain ?ob))
  :effect (and (harmony) (province ?ob) (craves ?ob ?underob)
               (not (province ?underob)) (not (pain ?ob))))

(:action feast
  :parameters  (?ob ?underob)
  :precondition (and (craves ?ob ?underob) (province ?ob) (harmony))
  :effect (and (pain ?ob) (province ?underob)
               (not (craves ?ob ?underob)) (not (province ?ob)) (not (harmony)))))

\end{lstlisting}
\begin{lstlisting}[language=lisp, caption={Problem File for Mystery\_BlocksWorld-100}, label={lst:mystery_blocksworld_pf}]
(define (problem mystery_blocksworld-p01)
  (:domain mystery_blocksworld)
  (:objects a b c d )
  (:init 
    (craves a b)
    (craves b c)
    (harmony)
    (planet c)
    (planet d)
    (province a)
    (province d)
  )
  (:goal (and 
    (craves a d)
    (craves c a)
  ))
)
    \end{lstlisting}
\clearpage
    \begin{lstlisting}[language=lisp, caption={Domain Description for Mystery\_BlocksWorld-100}, label={lst:mystery_blocksworld_dd}]
<@\textcolor{OliveGreen}{I am playing with a set of objects. Here are the actions I can do}@>

   <@\textcolor{OliveGreen}{Attack object}@> 
   <@\textcolor{OliveGreen}{Feast object from another object}@>
   <@\textcolor{OliveGreen}{Succumb object}@>
   <@\textcolor{OliveGreen}{Overcome object from another object}@> 

I have the following restrictions on my actions: 
    To perform Attack action, the following facts need to be <@\textcolor{red}{true: Province object, Planet object, }@> 
    <@\textcolor{red}{Harmony.}@>
    Once Attack action is performed the following facts will be <@\textcolor{red}{true: Pain object.}@>
    Once Attack action is performed the following facts will be <@\textcolor{red}{false: Province object, Planet object, }@>
    <@\textcolor{red}{Harmony.}@>
    To perform Succumb action, the following facts need to be <@\textcolor{red}{true: Pain object.}@>
    Once Succumb action is performed the following facts will be <@\textcolor{red}{true: Province object, Planet object, }@>
    <@\textcolor{red}{Harmony.}@>
    Once Succumb action is performed the following facts will be <@\textcolor{red}{false: Pain object.}@>
    To perform Overcome action, the following needs to be <@\textcolor{red}{true: Province other object, Pain object.}@>
    Once Overcome action is performed the following will be <@\textcolor{red}{true: Harmony, Province object, }@>
    <@\textcolor{red}{Object Craves other object.}@>
    Once Overcome action is performed the following will be <@\textcolor{red}{false: Province other object, Pain object.}@>
    To perform Feast action, the following needs to be <@\textcolor{red}{true: Object Craves other object, Province }@>
    <@\textcolor{red}{object, Harmony.}@>
    Once Feast action is performed the following will be <@\textcolor{red}{true: Pain object, Province other object.}@>
    Once Feast action is performed the following will be <@\textcolor{red}{false:, Object Craves other object, Province }@>
    <@\textcolor{red}{object, Harmony.}@>
    \end{lstlisting}

    \begin{lstlisting}[language=lisp, caption={Problem Description for Mystery\_BlocksWorld-100}, label={lst:mystery_blocksworld_pd}]
As initial conditions I have that, <@\textcolor{OliveGreen}{object a craves object b, object b craves object }@><@\textcolor{OliveGreen}{c}@>, harmony, 
<@\textcolor{red}{planet object c, planet object d, province object a and province object d}@>. 
My goal is to have that object a craves object d and object c craves object a.
    \end{lstlisting}

    \begin{lstlisting}[language=lisp, caption={Prompt for LLM-as-Planner on BlocksWorld-100}, label={lst:blocksworld_planner}]
Here is a game we are playing.

{domain_description}

{problem_description}

Write the plan that would solve this
problem.
These are the available actions:
(PICK-UP block): pick up a block from the
table
(PUT-DOWN block): put down a block on the
table
(STACK block1 block2): stack block1 onto
block2
(UNSTACK block1 block2): unstack block1
from block2

Here is what the output should look like:
(PICK-UP A)
(STACK A B)
(UNSTACK A B)
(PUT-DOWN A)
    \end{lstlisting}
\clearpage
    \begin{lstlisting}[language=lisp, caption={Prompt for LLM-as-Planner on Mystery\_BlocksWorld-100}, label={lst:mystery_blocksworld_planner}]
Here is a game we are playing.

{domain_description}

{problem_description}

Write the plan that would solve this
problem.
These are the available actions:
(ATTACK object): attack object
(SUCCUMB object): succumb
(OVERCOME object1 object2): overcome object1 from object2
(FEAST object1 object2): feast object1 from object2

Here is what the output should look like:
(ATTACK A)
(OVERCOME A B)
(FEAST A B)
(SUCCUMB A)
    \end{lstlisting}

    \begin{lstlisting}[language=lisp, caption={Prompt for LLM-as-Planner on Logistics-100}, label={lst:logistics_planner}]
Here is a game we are playing.

{domain_description}

{problem_description}

Write the plan that would solve this
problem.
These are the available actions:
(load-truck package truck location): load a package onto a truck at a location
(load-airplane object airplane location): load a package onto an airplane at a location
(unload-truck package truck location): unload a package from a truck at a location
(unload-airplane package airplane location): unload a package from an airplan at a location
(drive-truck truck location1 location2 city): drive a truck from location1 to location2 in a city
(fly-airplane airplane location1 location2): fly an airplane from location1 to location2

Here is what the output should look like:
(load-truck package truck city1-1)
(drive-truck truck city1-1 city1-2 city1)
(unload-truck package truck city1-2)
(load-airplane package plane city1-2)
(fly-airplane plane city1-2 city2-1)
(unload-airplane package plane city2-1)
    \end{lstlisting}
\clearpage
    \begin{lstlisting}[language=lisp, caption={Prompt for LLM-as-Planner on Barman-100}, label={lst:barman_planner}]
Here is a game we are playing.

{domain_description}

{problem_description}

Write the plan that would solve this
problem.
These are the available actions:
(GRASP hand container): grasp container with hand
(LEAVE hand container): leave container with hand
(FILL-SHOT shot ingredient hand1 hand2 dispenser): fill shot with ingredient from dispenser using hand1 and hand2
(REFILL-SHOT shot ingredient hand1 hand2 dispenser): re-fill shot with ingredient from dispenser using hand1 and hand2
(EMPTY-SHOT hand shot beverage): empty shot containing beverage with hand
(CLEAN-SHOT shot beverage hand1 hand2): clean shot containing beverage using hand1 and hand2
(POUR-SHOT-TO-CLEAN-SHAKER shot ingredient shaker hand level level1): pour shot containing ingredient into clean shaker using hand so that the level in the shaker changes from level to level1
(POUR-SHOT-TO-USED-SHAKER shot ingredient shaker hand level level1): pour shot containing ingredient into used shaker using hand so that the level in the shaker changes from level to level1
(EMPTY-SHAKER hand shaker cocktail level level1): empty shaker containing cocktail with hand so that the level in the shaker changes from level to level1
(CLEAN-SHAKER hand1 hand2 shaker): clean shaker using hand1 and hand2
(SHAKE cocktail ingredient1 ingredient2 shaker hand1 hand2): shake shaker containing cocktail of ingredient1 and ingredient2 using hand1 and hand2
(POUR-SHAKER-TO-SHOT beverage shot hand shaker level level1): pour shaker containing beverage into shot using hand so that the level in the shaker changes from level to level1

Here is what the output should look like:
(GRASP right shot1)
(FILL-SHOT shot1 ingredient10 right left dispenser10)
(POUR-SHOT-TO-CLEAN-SHAKER shot1 ingredient10 shaker1 right l0 l1)
(CLEAN-SHOT shot1 ingredient10 right left)
(FILL-SHOT shot1 ingredient5 right left dispenser5)
(GRASP left shaker1)
(POUR-SHOT-TO-USED-SHAKER shot1 ingredient5 shaker1 right l1 l2)
(LEAVE right shot1)
(SHAKE cocktail1 ingredient5 ingredient10 shaker1 left right)
(LEAVE left shaker1)
(GRASP left shot1)
(CLEAN-SHOT shot1 ingredient5 left right)
(GRASP right shaker1)
(POUR-SHAKER-TO-SHOT cocktail1 shot1 right shaker1 l2 l1)
    \end{lstlisting}
\twocolumn

\begin{table*}[!t]
\centering
\small
\resizebox{\textwidth}{!}{
\begin{tabular}{lllllll}
\toprule
\multicolumn{7}{c}{Metrics} \\ \midrule
 & \multicolumn{2}{c}{Natural} & \multicolumn{2}{c}{Moderately Templated} & \multicolumn{2}{c}{Heavily Templated}\\
Models & Solvability & Correctness & Solvability & Correctness & Solvability & Correctness\\ \midrule
\texttt{gemma-2-9b-it} & 3/100 & 3/100 & 0/100 & 0/100 & 61/100 & 10/100\\
\texttt{gemma-2-9b-it$^{p}$} & - & 9/100 & - & 5/100 & - & 5/100\\
\texttt{gemma-2-27b-it} & 0/100 & 0/100 & 17/100 & 10/100 & 81/100 & 80/100\\
\texttt{gemma-2-27b-it$^{p}$} & - & 11/100 & - & 3/100 & - & 7/100 \\ \midrule
\texttt{Llama-3.1-8B} & 0/100 & 0/100 & 0/100 & 0/100 & 0/100 & 0/100\\
\texttt{Llama-3.1-8B$^{p}$} & - & 1/100 & - & 1/100 & - & 0/100 \\
\texttt{Llama-3.1-70B} & 0/100 & 0/100 & 0/100 & 0/100 & 0/100 & 0/100\\
\texttt{Llama-3.1-70B$^{p}$} & - & 13/100 & - & 10/100 & - & 10/100\\ \midrule
\texttt{DeepSeek-R1-Distill-Llama-8B} & 0/100 & 0/100 & 0/100 & 0/100 & 0/100 & 0/100 \\
\texttt{DeepSeek-R1-Distill-Llama-8B$^{p}$} & - & 2/100 & - & 2/100 & - & 2/100 \\
\texttt{DeepSeek-R1-Distill-Llama-70B} & 0/100 & 0/100 & 0/100 & 0/100 & 0/100 & 0/100\\
\texttt{DeepSeek-R1-Distill-Llama-70B$^{p}$} & - & 18/100 & - & 25/100 & - & 22/100 \\
\texttt{DeepSeek-R1} & 83/100 & 81/100 & 76/100 & 75/100 & 84/100 & 84/100\\
\texttt{DeepSeek-R1$^{p}$} & - & 91/100 & - & 85/100 & - & 86/100 \\
\midrule
\texttt{gpt-3.5-turbo} & 2/100 & 1/100 & 14/100 & 4/100 & 39/100 & 29/100 \\
\texttt{gpt-4o-mini} & 19/100 & 3/100 & 9/100 & 5/100 & 66/100 & 59/100 \\ 
\texttt{gpt-4o-mini$^{p}$} & - & 7/100 & - & 7/100 & - & 1/100 \\
\texttt{gpt-4o} & 64/100 & 60/100 & 77/100 & 67/100 & 89/100 & 89/100 \\ 
\texttt{gpt-4o$^{p}$} & - & 33/100 & - & 35/100 & - & 29/100 \\
\texttt{o1-preview} & 91/100 & 91/100 & - & - & - & - \\
\texttt{o1-preview$^{p}$} & - & 82/100  & - & - & - & - \\
\texttt{o3-mini} & 79/100 & 68/100 & 82/100 & 70/100 & 94/100 & 94/100 \\
\texttt{o3-mini$^{p}$} & - & 87/100 & - & 87/100 & - & 96/100 \\
\bottomrule

\end{tabular}
}
\caption{Performance of LLM-as-formalizer and LLM-as-planner ($^{p}$) on all naturalness settings of the \bw dataset. }
    \label{tab:BW-100}
\end{table*}

%\hc{Table 1 equivalent for the 2 templated versions.}

\begin{table*}[!t]
    \centering
    \small
    \begin{tabular}{lll} \toprule
     %& \multicolumn{2}{c}{Metrics} \\ \midrule
      Models & Solvability & Correctness \\ \midrule
      \texttt{Llama-3.1-8B} & 0/100 & 0/100 \\
      \texttt{Llama-3.1-8B$^{p}$} & - & 0/100 \\
      \texttt{Llama-3.1-70B} & 0/100 & 0/100 \\
      \texttt{Llama-3.1-70B$^{p}$} & - & 0/100 \\ \midrule
      \texttt{gemma-2-9b-it} & 100/100 & 99/100 \\
      \texttt{gemma-2-9b-it$^{p}$} & - & 9/100 \\
      \texttt{gemma-2-27b-it} & 99/100 & 98/100 \\
      \texttt{gemma-2-27b-it$^{p}$} & - & 0/100 \\ \midrule
      \texttt{DeepSeek-R1} & 56/100 & 54/100 \\
      \texttt{DeepSeek-R1$^{p}$} & - & 21/100 \\ \midrule
     \texttt{gpt-3.5-turbo} & 4/100 & 0/100 \\
     \texttt{gpt-4o-mini} & 36/100 & 5/100 \\ 
     \texttt{gpt-4o-mini$^{p}$} & - & 0/100 \\
     \texttt{gpt-4o} & 74/100 & 70/100 \\ 
     \texttt{gpt-4o$^{p}$} & - & 0/100 \\
     \texttt{o3-mini} & 95/100 & 95/100 \\ 
     \texttt{o3-mini$^{p}$} & - & 74/100 \\
\bottomrule     
    \end{tabular}
    \caption{Performance of LLM-as-formalizer and LLM-as-planner ($^{p}$) on the Heavily Templated \mbw dataset. }
    \label{tab:Heavily Templated MBW-111}
\end{table*}

\begin{table*}[!t]
\centering
\small
\begin{tabular}{lllllll}
\toprule
\multicolumn{7}{c}{Metrics} \\ \midrule
 & \multicolumn{2}{c}{Natural} & \multicolumn{2}{c}{Moderately Templated} & \multicolumn{2}{c}{Heavily Templated} \\
Models & Solvability & Correctness & Solvability & Correctness & Solvability & Correctness \\ \midrule
\texttt{gemma-2-9b-it} & 1/100 & 0/100 & 0/100 & 0/100 & 1/100 & 0/100 \\
\texttt{gemma-2-9b-it$^{p}$} & - & 0/100 & - & 0/100 & - & 0/100 \\
\texttt{gemma-2-27b-it} & 1/100 & 0/100 & 3/100 & 0/100 & 0/100 & 0/100 \\
\texttt{gemma-2-27b-it$^{p}$} & - & 0/100 & - & 0/100 & - & 1/100 \\ \midrule
\texttt{Llama-3.1-8B} & 0/100 & 0/100 & 0/100 & 0/100 & 0/100 & 0/100 \\
\texttt{Llama-3.1-8B$^{p}$} & - & 1/100 & - & 0/100 & - & 0/100 \\
\texttt{Llama-3.1-70B} & 0/100 & 0/100 & 0/100 & 0/100 & 0/100 & 0/100 \\
\texttt{Llama-3.1-70B$^{p}$} & - & 0/100 & - & 0/100 & - & 0/100 \\ \midrule
\texttt{DeepSeek-R1} & 51/100 & 9/100 & 73/100 & 73/100 & 64/100 & 64/100 \\
\texttt{DeepSeek-R1$^{p}$} & - & 12/100 & - & 22/100 & - & 22/100 \\ \midrule
\texttt{gpt-3.5-turbo} & 1/100 & 0/100 & 0/100 & 0/100 & 1/100 & 1/100 \\
\texttt{gpt-3.5-turbo}$^{p}$ & - & 0/100 & - & 0/100 & - & 0/100 \\
\texttt{gpt-4o-mini} & 13/100 & 0/100 & 29/100 & 0/100 & 26/100 & 0/100 \\ 
\texttt{gpt-4o-mini$^{p}$} & - & 0/100 & - & 0/100 & - & 0/100 \\
\texttt{gpt-4o} & 20/100 & 2/100 & 34/100 & 13/100 & 33/100 & 13/100 \\ 
\texttt{gpt-4o$^{p}$} & - & 1/100 & - & 0/100 & - & 0/100 \\
\texttt{o3-mini} & 43/100 & 7/100 & 51/100 & 39/100 & 55/100 & 47/100 \\
\texttt{o3-mini$^{p}$} & - & 14/100 & - & 21/100 & - & 21/100 \\ 
\bottomrule
\end{tabular}
\caption{Performance of LLM-as-formalizer and LLM-as-planner ($^{p}$) on all \lgt datasets.}
\label{tab:logistics}
\end{table*}

\begin{table*}[t!]
    \centering
    \small
    \begin{tabular}{lll} \toprule
     %& \multicolumn{2}{c}{Metrics} \\ \midrule
      Models & Solvability & Correctness \\ \midrule
      \texttt{Llama-3.1-8B} & 0/100 & 0/100 \\
      \texttt{Llama-3.1-8B$^{p}$} & - & 0/100 \\
      \texttt{Llama-3.1-70B} & 0/100 & 0/100 \\
      \texttt{Llama-3.1-70B$^{p}$} & - & 15/100 \\ \midrule
      \texttt{gemma-2-9b-it} & 0/100 & 0/100 \\
      \texttt{gemma-2-9b-it$^{p}$} & - & 2/100 \\
      \texttt{gemma-2-27b-it} & 0/100 & 0/100 \\
      \texttt{gemma-2-27b-it$^{p}$} & - & 0/100 \\ \midrule
      \texttt{DeepSeek-R1} & 10/100 & 2/100 \\
      \texttt{DeepSeek-R1$^{p}$} & - & 22/100 \\ \midrule
     \texttt{gpt-3.5-turbo} & 0/100 & 0/100 \\
     \texttt{gpt-3.5-turbo$^{p}$} & - & 4/100 \\
     \texttt{gpt-4o-mini} & 0/100 & 0/100 \\ 
     \texttt{gpt-4o-mini$^{p}$} & - & 8/100 \\
     \texttt{gpt-4o} & 1/100 & 0/100 \\ 
     \texttt{gpt-4o$^{p}$} & - & 3/100 \\
     \texttt{o3-mini} & 1/100 & 0/100 \\ 
     \texttt{o3-mini$^{p}$} & - &  19/100\\
\bottomrule     
    \end{tabular}
    \caption{Performance of LLM-as-formalizer and LLM-as-planner ($^{p}$) on the Heavily Templated \bm dataset. }
    \label{tab:barman}
\end{table*}

\begin{table*}[!t]
\centering
\small
\begin{tabular}{lllll}
\toprule
\multicolumn{5}{c}{Metrics} \\ \midrule
 & \multicolumn{2}{c}{Natural} & \multicolumn{2}{c}{Heavily Templated} \\
Models & Solvability & Correctness & Solvability & Correctness \\ \midrule
\texttt{Llama-3.1-8B} & 1/100 & 1/100 & 49/100 & 45/100 \\
\texttt{Llama-3.1-70B} & 0/100 & 0/100 & 0/100 & 0/100 \\ \midrule
\texttt{gpt-3.5-turbo} & 8/100 & 1/100 & 35/100 & 24/100 \\
\texttt{gpt-4o-mini} & 13/100 & 3/100 & 75/100 & 61/100 \\ 
\texttt{gpt-4o} & 60/100 & 48/100 & 93/100 & 92/100 \\ 
\bottomrule
\end{tabular}
\caption{Performance of LLM-as-formalizer all the Natural and Heavily Templated \bw datasets using few-shot prompting.}
\label{tab:bw_few-shot}
\end{table*}

\begin{table*}[!t]
\centering
\small
\begin{tabular}{lllll}
\toprule
\multicolumn{5}{c}{Metrics} \\ \midrule
 & \multicolumn{2}{c}{Natural} & \multicolumn{2}{c}{Heavily Templated} \\
Models & Solvability & Correctness & Solvability & Correctness \\ \midrule
\texttt{Llama-3.1-8B} & 0/100 & 0/100 & 0/100 & 0/100 \\
\texttt{Llama-3.1-70B} & 0/100 & 0/100 & 0/100 & 0/100 \\ \midrule
\texttt{gpt-3.5-turbo} & 6/100 & 0/100 & 23/100 & 17/100 \\
\texttt{gpt-4o-mini} & 19/100 & 4/100 & 49/100 & 45/100 \\ 
\texttt{gpt-4o} & 61/100 & 48/100 & 91/100 & 90/100 \\ 
\bottomrule
\end{tabular}
\caption{Performance of LLM-as-formalizer all the Natural and Heavily Templated \bw datasets using chain-of-thought prompting.}
\label{tab:bw_cot}
\end{table*}

\begin{table*}[t!]
\centering
\small
\resizebox{\columnwidth}{!}{\begin{tabular}{llll}
\toprule
     Models & Syntax Error & \df Error & \pf Error\\ \midrule
     & \multicolumn{3}{c}{Natural \bw} \\
     \texttt{gemma-2-9b-it} & 15/20 & 20/20 & 20/20 \\ 
     \texttt{gemma-2-27b-it} & 3/20 & 20/20 & 14/20 \\
     \texttt{Llama-3.1-8B} & 20/20 & 20/20 & 18/20 \\
     \texttt{Llama-3.1-70B} & 20/20 & 20/20 & 17/20 \\
     %\texttt{gpt-3.5-turbo} & 10/20 & 20/20 & 20/20 \\
     \texttt{gpt-4o-mini} & 2/20 & 20/20 & 19/20 \\
     \texttt{gpt-4o} & 2/20 & 2/20 & 18/20 \\  \midrule
     & \multicolumn{3}{c}{Natural \lgt} \\
     \texttt{gemma-2-9b-it} & 7/20 & 20/20 & 15/20 \\ 
     \texttt{gemma-2-27b-it} & 8/20 & 20/20 & 20/20 \\
     \texttt{Llama-3.1-8B} & 20/20 & 20/20 & 20/20 \\ 
     \texttt{Llama-3.1-70B} & 20/20 & 20/20 & 10/20 \\
     %\texttt{gpt-3.5-turbo} & 10/20 & 20/20 & 20/20 \\
     \texttt{gpt-4o-mini} & 2/20 & 20/20 &  19/20\\ 
     \texttt{gpt-4o} & 5/20 & 20/20 & 19/20 \\  \midrule
     & \multicolumn{3}{c}{\mbw}\\
     %\texttt{gpt-3.5-turbo} & 16/20 & 19/20 & 2/20 \\ 
     \texttt{gpt-4o-mini} & 6/20 & 20/20 & 1/20 \\ 
     \texttt{gpt-4o} & 5/20 & 16/20 & 0/20 \\      
\bottomrule
\end{tabular}}
\caption{Annotated types of 20 randomly sampled errors of LLM-as-formalizer for various models and datasets.}
\label{tab:analysis of errors}
\end{table*}

\begin{table*}[t!]
    \centering
    \resizebox{\textwidth}{!}{
    \small
    \begin{tabular}{llllll}
    \toprule
     Models & Wrong Precondition & Wrong Effect & Missing Predicate & Missing Action & Missing Parameters \\ \midrule
     & \multicolumn{5}{c}{Natural \bw} \\ \midrule
     gpt-4o-mini & 11/20 & 18/20 & 19/20 & 1/20 & 2/20 \\
     gpt-4o & 0/20 & 2/20 & 0/20 & 0/20 & 0/20 \\
    \midrule
    & \multicolumn{5}{c}{Natural \lgt} \\
     \midrule
     gpt-4o-mini & 20/20 & 16/20 & 20/20 & 5/20 & 17/20 \\
     gpt-4o & 17/20 & 9/20 & 19/20 &1/20  & 9/20 \\
    \midrule
    & \multicolumn{5}{c}{Heavily Templated \mbw} \\
     \midrule
     gpt-4o-mini & 14/20 & 17/20 & 17/20 & 0/20 & 5/20 \\
     gpt-4o & 13/20 & 14/20 & 0/20 & 0/20 & 2/20 \\
     \bottomrule
    \end{tabular}
    }
    \caption{Analysis of errors found in \df for Natural \bw and Heavily Templated \mbw out of 20 randomly sampled instances.}
    \label{tab:analysis of df errors}
\end{table*}

\clearpage
\onecolumn
    \begin{lstlisting}[language=lisp, caption={Example Domain Description and Problem Description given to Llama-3.1-8B-Instruct}, label={lst:llama_dd_pd}]
;Domain Description:
Blocksworld is a puzzle domain where the aim is to move a set of blocks to achieve a specific goal configuration. The domain consists of a flat surface, often called the 'table', and an arbitrary number of blocks. Every block can be stacked on top of another block or placed on the table. Blocks can be moved one at a time, and only the block at the top of a stack can be moved. The main actions available in Blocksworld are: 


1. **Pick Up (Pick-Up B)**: Picks up block B from the table.
2. **Put Down (Put-Down B)**: Places block B onto the table. 
3. **Stack (Stack B1 B2)**: Stacks block B1 on top of block B2.
4. **Unstack (Unstack B1 B2)**: Removes block B1 from the top of block B2. 

The goal is to transform an initial configuration of blocks into a desired target configuration using these actions. 

;Problem Description:
In this Blocksworld problem, we will consider a simple scenario with 2 blocks: Block1 and Block2. The initial state of the blocks is as follows: 

**Initial State**: 
- Block1 is on the table. 
- Block2 is on Block1. 

The goal is to achieve the following configuration: 

**Goal State**: 
- Block2 is on the table. 
- Block1 is on the table. 

To solve this problem, you will need to use the available actions to rearrange the blocks from the initial state to the goal state.       
    \end{lstlisting}

    \begin{lstlisting}[language=lisp, caption={Generated Domain File from Llama-3.1-8B-Instruct}, label={lst:llama_df}]
(define (domain blocksworld)
  (:requirements :strips :typing)
  (:types block)
  <@\textcolor{red}{(:predicates (on-table?b - block) (on?b1 - block?b2 - block) (clear?b - block) ) }@>
  (:action pick-up 
    :parameters (?b - block) 
    <@\textcolor{red}{:preconditions}@> (and (on-table?b) (clear?b)) 
    :effects (and (not (on-table?b)) (not (clear?b)) <@\textcolor{red}{(on-table?b))} @>
  ) 
  (:action put-down
    :parameters (?b - block)
    <@\textcolor{red}{:preconditions}@> (and (not (on-table?b)) <@\textcolor{red}{(clear?b))}@>
    :effects (and (on-table?b) (clear?b))
  )
  (:action stack
    :parameters (?b1 - block?b2 - block)
    <@\textcolor{red}{:preconditions}@> (and <@\textcolor{red}{(on-table?b1)}@> (clear?b2))
    :effects (and (not (on-table?b1)) (not (clear?b2)) (on?b1?b2))
  )
  (:action unstack
    :parameters (?b1 - block?b2 - block)
    <@\textcolor{red}{:preconditions}@> (and (on?b1?b2) (clear?b1))
    :effects (and (on-table?b1) (not (on?b1?b2)) (clear?b2))
  )
)
\end{lstlisting}
\clearpage
\begin{lstlisting}[language=lisp, caption={Generated Problem File from Llama-3.1-8B-Instruct}, label={lst:llama_pf}]
(define (problem blocksworld-problem)
  (:domain blocksworld)
  (:objects block1 block2 - block)
  (:init
    (on-table block1)
    <@\textcolor{red}{(on block2 block1)}@>
    <@\textcolor{red}{(clear block1)} @>
    (clear block2)
  )
  (:goal
    (and (on-table block2)
         (on-table block1)
         (clear block1)
         (clear block2)
    )
  )
)
    \end{lstlisting}

\twocolumn

\iffalse
\begin{figure*}
    \centering
    \begin{adjustbox}{max size={\textwidth}{\textheight}}
        \includegraphics[]{figures/pddl_example.png}
    \end{adjustbox}
    \caption{Example of PDDL generated using \texttt{Meta-Llama-3.1-8B-Instruct}. We can see that there are syntax errors due to using the incorrect keyword "preconditions" instead of "precondition". However, there are more errors such as incorrect predicates in the initial state of the \pf, and incorrect predicates, preconditions and effects in the \df. Examples of errors are highlighted in red text.}
    \label{fig: example pddl}
\end{figure*}
\fi

\end{document}